\documentclass[runningheads]{llncs}
\usepackage{graphicx}
\usepackage{amsmath,amssymb} % define this before the line numbering.
\usepackage{color}
\usepackage[width=122mm,left=12mm,paperwidth=146mm,height=193mm,top=12mm,paperheight=217mm]{geometry}
\usepackage{times}
\usepackage{epsfig}
\usepackage{graphicx}
\usepackage{epstopdf}
\usepackage{amsmath}
\usepackage{amssymb}
\usepackage{bm}
\usepackage{subfigure}
\usepackage{tabularx}
\usepackage{algorithm}
\usepackage{algorithmicx}
\usepackage{algpseudocode}
\usepackage{multirow}

\begin{document}
 %\renewcommand\makeLineNumber {\hss\thelinenumber\ \hspace{6mm} \rlap{\hskip\textwidth\ \hspace{6.5mm}\thelinenumber}}
%\linenumbers
\pagestyle{headings}
\mainmatter
\def\ECCV18SubNumber{667}  % Insert your submission number here

\title{Region-filtering Correlation Tracking} % Replace with your title

%\titlerunning{ECCV-18 submission ID \ECCV18SubNumber}

%\authorrunning{ECCV-18 submission ID \ECCV18SubNumber}

%\author{Anonymous ECCV submission}
%\institute{Paper ID \ECCV18SubNumber}
\author{Nana Fan, Zhenyu He}
\institute{Harbin Institute of Technology}

\maketitle

\begin{abstract}
Recently, correlation filters have demonstrated the excellent performance in visual tracking.
However, the base training sample region is larger than the object region, including the Interference Region (IR). The IRs in training samples from cyclic shifts of the base training sample severely degrade the quality of a tracking model.
%They not only contain a object, but also include the surrounding of the object.
% 我们提出了一个扩展版本
In this paper, we propose the novel Region-filtering Correlation Tracking (RFCT) to address this problem. We immediately filter training samples by introducing a spatial map into the standard CF formulation.
%and present an optimization strategy to learn a correlation filter.
Compared with existing correlation filter trackers, our proposed tracker has the following advantages:
(1)
%A spatial map is able to eliminate the interference of IRs in training samples, by introducing the spatial map into standard CF formulation.
The correlation filter can be learned on a larger search region without the interference of the IR by a spatial map.
(2)
%Because of a spatial map directly processing training samples, the values of the spatial map are not restricted.
Due to processing training samples by a spatial map, it is more general way to control background information and target information in training samples. The values of the spatial map are not restricted, then a better spatial map can be explored.
(3) The weight proportions of accurate filters are increased to alleviate model corruption.
Experiments are performed on two benchmark datasets: OTB-2013 and OTB-2015. Quantitative evaluations on these benchmarks demonstrate that the proposed RFCT algorithm performs favorably against several state-of-the-art methods.
\keywords{Visual Tracking, Correlation Filter}
\end{abstract}

\section{Introduction}

Visual tracking is a fundamental problem in computer vision. Given the position of an object in the first frame, visual tracking aims to sequentially locate the object in the rest of an image sequence. Although many algorithms have been proposed in the recent years, visual tracking is still a challenge as a result of complex scenes.

In recent years, correlation filter based methods \cite{danelljan2016eco,danelljan2016beyond,kiani2017learning,mueller2017context} have shown notable performance in standard benchmarks, partly because of their efficiency in exploiting more training samples.
% correlation filter  has wide attention, because of its efficiently exploiting more training samples.
The correlation filter is introduced by Bolme \emph{et al.} \cite{bolme2010visual} into visual tracking, correlated over an exemplar to gain a desired Gaussian response. It is time-consuming to immediately derive the correlation filter from least square loss in the time domain. Therefore, to create a fast tracker, the correlation operation is transformed to the Fourier domain
 %Fast Fourier Transform \cite{press1989numerical}
 by Convolution Theorem. Then, the correlation filter can be solved by the efficiently element-wise operation. KCF \cite{henriques2015high} explains the correlation filter by means of generally known ridge regression and derives a new kernelized correlation filter. In this way, the correlation is reformulated as that the circulant matrix, which denotes all training samples from circulating shift of a base training sample, multiplies by the filter. It provides a deeper understanding for the correlation filter. And the result is the same as that in \cite{bolme2010visual}. Correlation filter trackers have achieved state-of-the-art performances, but they still have a problem in the training procedure.

 \begin{figure}
\centering
\subfigure[Original image]{\includegraphics[height=1.5in]{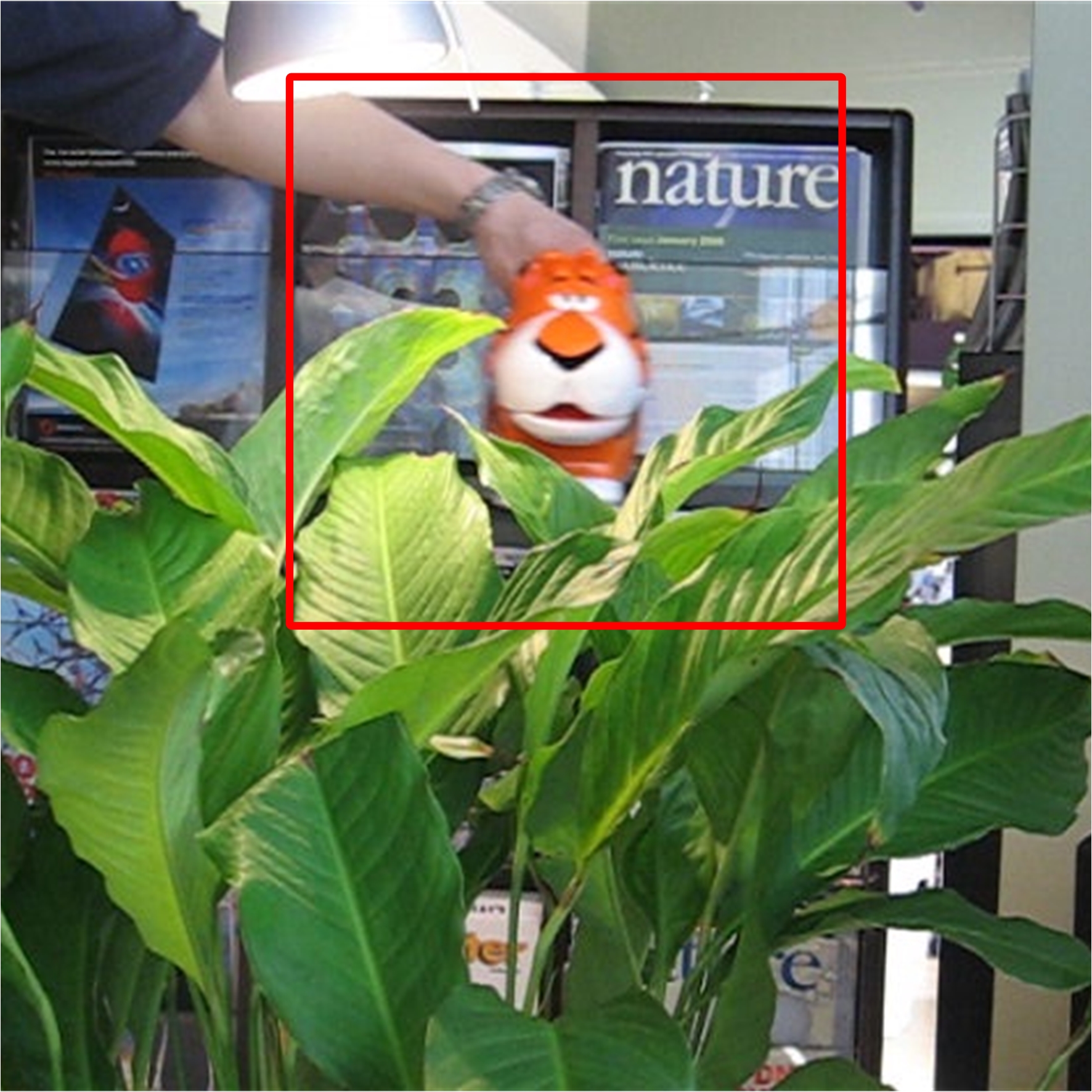}} \ \
\subfigure[Training samples]{\includegraphics[height=1.5in]{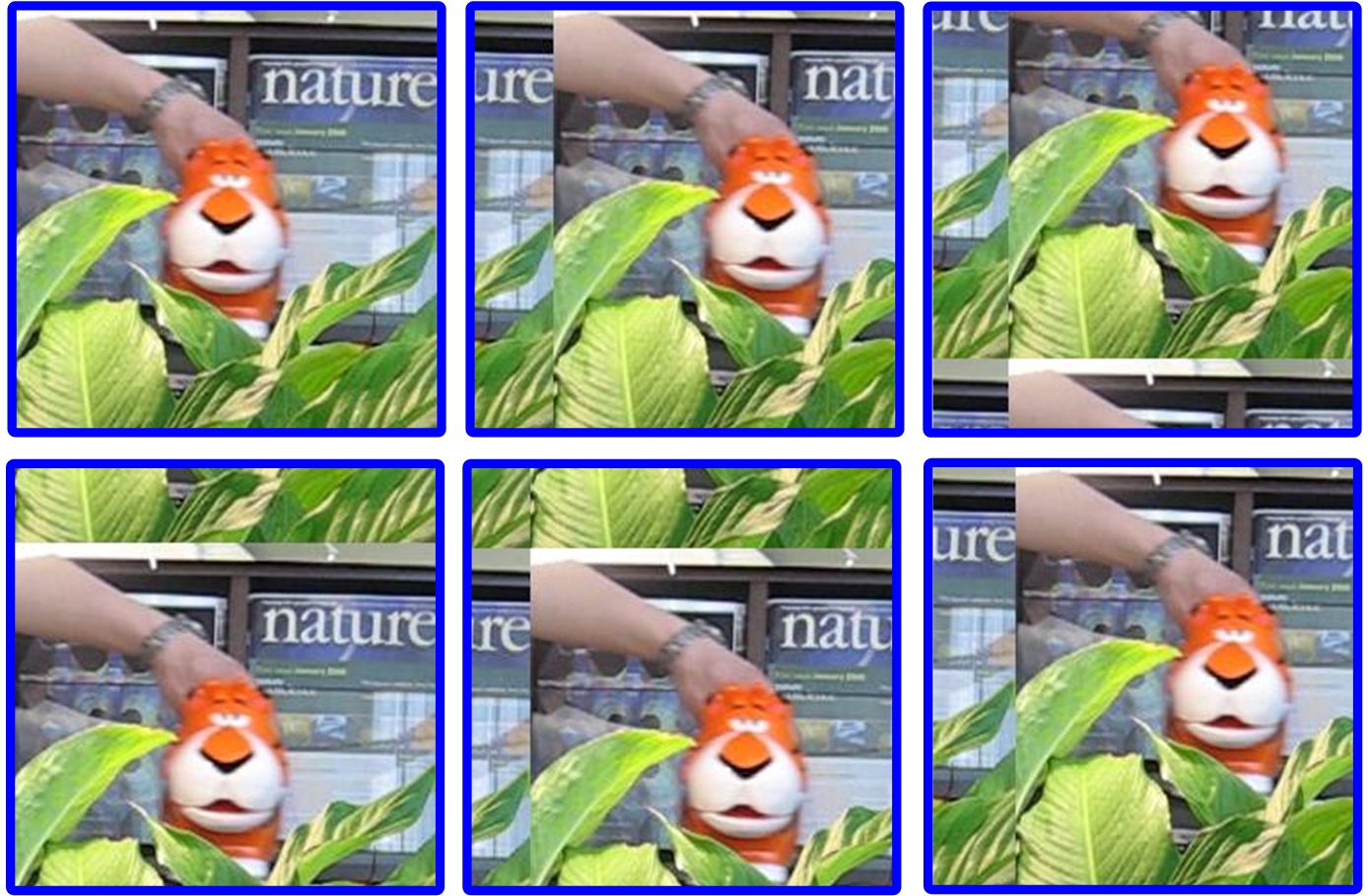}}
\vspace{-2mm}\caption{(a) An original image and a base training sample.  (b) Some training samples are from cyclic shifts of the base training sample. The base training sample region is larger than the object region, including the IR. It leads to that training samples contains IRs, which severely degrade the tracking quality.}
\label{Fig:Problem}
\end{figure}

~\\
\textbf{Problem in Training Procedure:} When the correlation filter is introduced into visual tracking, it needs to fit the tracking task.
%the tracking task on which to base the filter.
Generally, to detect the object by one correlation operation in the next frame, the base training sample region is larger than the target region (see Fig. \ref{Fig:Problem}(a)). Training samples are from cyclic shifts \cite{henriques2015high} of the base training sample (see Fig. \ref{Fig:Problem}(b)). By the Convolution Theorem, the correlation in the time domain corresponds to an element-wise multiplication in the Fourier domain. Therefore, the filter has the same size as the training samples. Obviously, the filter is not modeled for the object, but modeled for the object and its background in the base training sample. We name the background region Interference Region (IR). The IRs in training samples severely degrade the tracking quality.
%From the viewpoint of Ridge Regression, the filter is a regressor estimating a test or detecting sample. But, training samples or detecting samples not only include the object, but also contain the surrounding background of the object (see figure \ref{Fig:Problem}).
%And the larger the search region in detection is, the as a result of more surrounding background in the base training sample. It will more affect the resulting performance.
This is the challenge that this paper addresses.
% 此处应该有说明图

To deal with the above problem, SRDCF \cite{danelljan2015learning} introduces a regularization component to penalize correlation filter coefficients depending on the corresponding spatial location. But, it only can penalize the correlation filter coefficients which are far away from the center. When a search region is very large, the edge coefficients of the filter should not works. Different to CFBL \cite{kiani2015correlation} and BACF \cite{kiani2017learning}, we model a filter for the base training sample, then exploit a spatial map to process the training samples. It is a more general way to control background information and target information within the training samples.

~\\
\textbf{Contribution:} In this paper, we propose Region-filtering Correlation Tracking (RFCT). The contributions of this work are summarized as follows.
%\textbf{(1)} We model a filter for the base training sample and introduce a spatial map into the standard CF formulation for processing all the training samples. Compared with the correlation filter
%trackers, it is a more general way to control background information and target information in training samples. The values of the spatial map are not restricted, then a better spatial map can be explored.
%\textbf{(2)} We increase the weight proportions of accurate filters to alleviate model corruption.

\renewcommand{\labelitemi}{$-$}
\begin{itemize}
\item We model a filter for the base training sample and introduce a spatial map into the standard CF formulation for processing all the training samples. Compared with the correlation filter
trackers, it is a more general way to control background information and target information in training samples. The values of the spatial map are not restricted, then a better spatial map can be explored.
\item We increase the weight proportions of accurate filters for model update to alleviate model corruption.
\end{itemize}

\begin{figure*}
\centering
\includegraphics[height=2.5in]{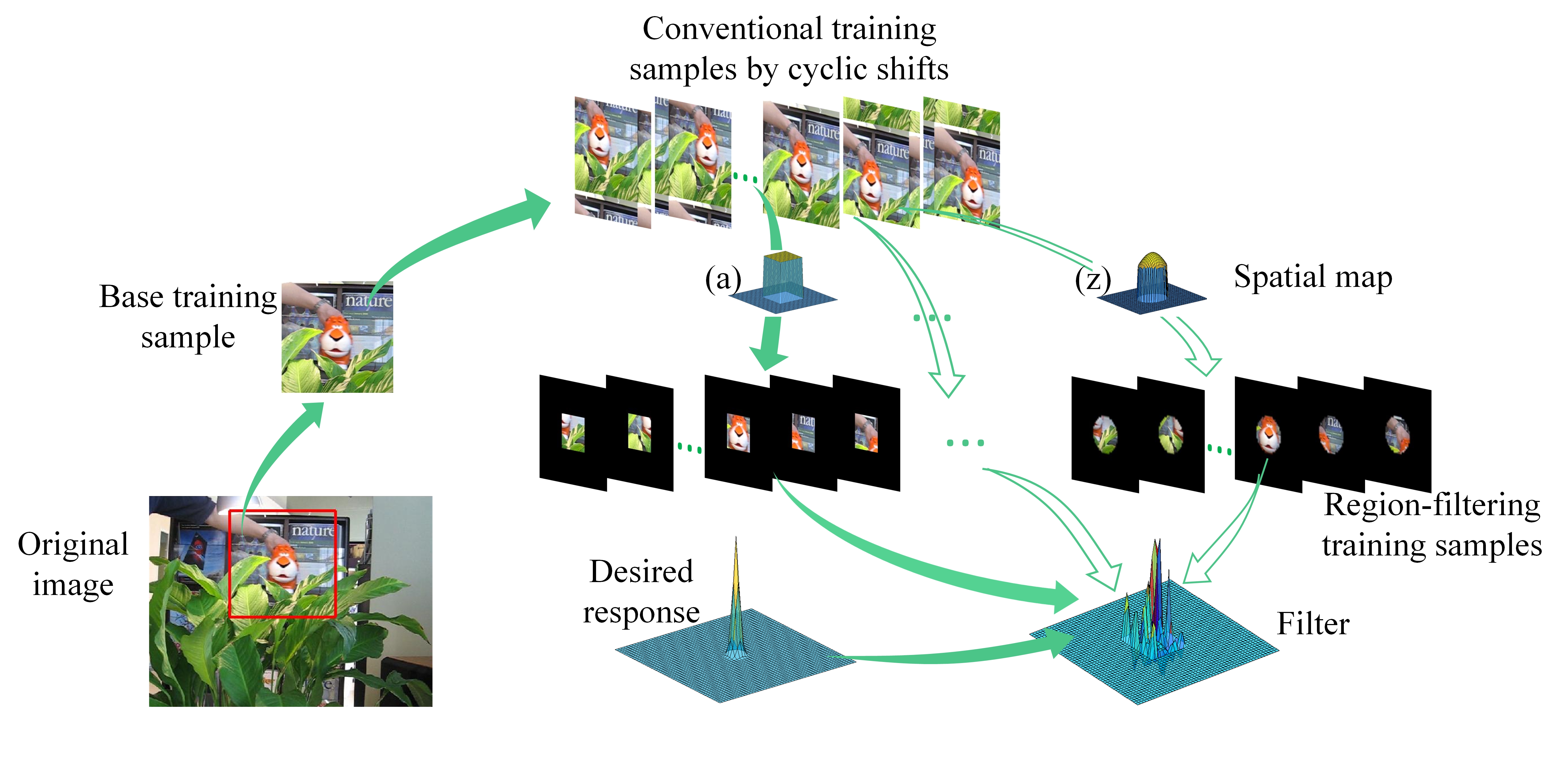}
\vspace{-9mm}\caption{Filtering the IRs. After locating the target, a base training sample is extracted, centered at the target. Training samples are from cyclic shifts of the base training sample. Because the base training sample region is larger than the target region, there are the IRs in training samples.
%the region of the base training sample is larger than the region of the target,
The IRs severely degrade the quality of the tracking algorithm. In our approach, a spatial map is used to filter training samples. The values of the spatial map are not restricted. For example, when a binary map (a) is used, it can eliminate the influence of the IR. Or when a map (z) is used, it can eliminate the influence of the IR and penalize the edge of the target region.}
\label{Fig:schematic}
\end{figure*}

\section{Related Work}
In visual tracking, the correlation filter has attracted wide attention, partly due to efficiently exploiting more training samples.
Bolme \emph{et al.} \cite{bolme2010visual} firstly introduce correlation filter into visual tracking with grayscale samples, keeping the object scale fixed in tracking
% no handling scale variations
. Afterwards, some methods \cite{kiani2013multi,danelljan2014adaptive,danelljan2015coloring,henriques2015high,li2014scale,tang2015multi} improve performance with the help of multi-channel features, such as HOG \cite{dalal2005histograms} or Color-Names \cite{van2009learning}.
% Building on it
KCF \cite{henriques2015high} and MKCF \cite{tang2015multi} make kernelized extensions of linear correlation filter to further improve the performance. On the basic of multi-channel features, DSST \cite{danelljan2014accurate} learns a correlation filter on a scale pyramid representation to catch the accurate object scale at real-time frame-rates. In term of the desired response, Bibi \emph{et al.} \cite{bibi2016target} offset the effect of using scores of circularly shifted samples, replacing the hand-crafted Gaussian response by an adaptive response with scores of actual translations.
%The other method,
Different from the traditional square loss, Wang \emph{et al.} \cite{wang2017large} draw lessons from the structured output tracker  \cite{hare2016struck} to learn a correlation filter in a large-margin framework. The other method \cite{sui2016real} proposes three sparse-based loss functions, and reveals the sensitivity of the peak values of the filter in successive frames is consistent with the tracking performance from an experimental perspective.
To deal with partial occlusion, some researchers \cite{liu2016structural,li2015reliable,liu2015real} have proposed to introduce the part-based strategy into correlation filter tracking. SCF \cite{liu2016structural} limits all individual parts to keep close to each other for preserving the target object structure. The two works \cite{li2015reliable,liu2015real} apply an individual correlation filter for each part and rely on more reliable parts to locate the object. In addition, there are also some methods \cite{zuo2016learning,zhang2016defense} with the aid of circulant matrix used in correlation filters to make the trackers run faster. Zuo \emph{et al.} \cite{zuo2016learning} reformulate the SVM model with circulant matrix expression and present an efficient alternating optimization method. CST \cite{zhang2016defense} introduces the circulant structure property into a sparse tracker to reduce particles and be solved efficiently.

%In correlation filter tracking, the search region is larger than the target region. After detecting, a base training sample is extracted from a region with the same size as the search region. And training samples are from cyclic shifts of the base training sample. Therefore, there are IRs in the training samples, which degrade the tracking quality. Galoogahi \emph{et al.} in CFBL \cite{kiani2015correlation} and BACF \cite{kiani2017learning} propose a mask matrix to crop new patches with the same size as the target from the training samples for training. And BACF is a multi-channel extension of CFBL. But spatial information of the new training samples is destroyed as a result of regarding the base sample as a vector.
%SRDCF \cite{danelljan2015learning} introduces a spatial regularization component to penalize correlation filter coefficients depending on their spatial location. However the correlation filter coefficients much far away from the center region should be completely constrained. CSR-DCF \cite{lukevzivc2016discriminative} immediately forces the correlation coefficients outside the target region to be zero and within the target region to be one. But the spatial map is restricted to a binary map. Different to the above algorithms, we immediately filter training samples from cyclic shifts of a base training sample by a spatial map with the same size as training samples, and maintain the spatial information of training samples. In addition, the values of the spatial map are not restricted, examples of the spatial map is shown in Figure \ref{Fig:schematic}.

In correlation filter tracking, a search region is larger than the target region. Therefore, the base training sample not only covers the target, but also contains the IR. Because training samples are from cyclic shifts of the base training sample, the IRs in the training samples degrade the tracking quality. To deal with the above problem, some works \cite{danelljan2015learning,lukevzivc2016discriminative,kiani2015correlation,kiani2017learning} have mode successful attempts. Different to them, we model a filter for a base training sample and exploit a spatial map to process all the training samples. The values of the spatial map are not restricted, examples of the spatial map are shown in Fig. \ref{Fig:schematic}. It is a general way to control background information and target information in the training samples.

\subsection{Standard Correlation Filter Tracking}
The aim of CF is to learn a correlation filter from a base training sample $x$ with $d$ channels. We indicate feature channel $l \in \left\{ 1,...,d\right\}$ of $x$ by $x_l$. All training samples from cyclic shifts of the base training sample have the same spatial size $M \times N$.
%with the base training sample $x$.
Then there is a training sample $x(m,n)$at each spatial location $(m,n) \in \Omega:=\left\{ 0,...,M-1\right\} \times \left\{ 0,...,N-1\right\}$. The desired response $y$ is a scalar function over $\Omega$, including a label for each location.
CF can be formulated in the spatial domain as a ridge regression,
\begin{equation}
\label{dcf_1}
\mathop{\mathrm{min}}_{w_l} \sum\limits_{l=1}^{d}||X_l^\top w_l-y||^2+\lambda||w_l||^2,
\end{equation}
where $X_l$ is the circulant matrix of the base training sample $x$ in the $l$-th channel, $\top$ denotes the transpose, $y$ is the desired response, $\lambda$ is a regularization parameter that controls overfitting and $w_l$ is the $l$-th channel of the filter $w$ with the same size as the base training sample $x$. Moreover, it can also be expressed by the convolution operation,
\begin{equation}
\label{dcf_2}
\mathop{\mathrm{min}}_{w_l} \sum\limits_{l=1}^{d}||x_l \ast w_l-y||^2+\lambda||w_l||^2.
\end{equation}
Here, $\ast$ denotes circular convolution. By Parseval's formula, Eq.\ref{dcf_2} can be transformed to the Fourier domain where the correlation filter is obtained by the element-wise operation.

Similar to the training stage, a test sample $z$ is used for detection by the convolution operation.
%correlated by the correlation filter in the Fourier domain for the detection.
The scores $S$ in the search region to which the test sample corresponds can be computed by the inverse Discrete Fourier Transform (DFT),
\begin{equation}
\label{dcf_3}
S(z) = \mathcal{F}^{-1}\left(\sum\limits_{l=1}^{d}\hat{z}_l\odot \hat{w}_l\right),
\end{equation}
where $\odot$ indicates element-wise product, $\hat{}$ is the DFT and $\mathcal{F}^{-1}$ denotes the inverse DFT. Generally, the location with the maximum in the search region is regarded as the tracking result.
%\begin{equation}
%\mathop{\mathrm{min}}_{\hat{w}} \sum\limits_{l=1}^{d}||\hat{x}_l \odot \hat{w}_l-y||^2+\lambda\sum\limits_{l=1}^{d}||\hat{w}_l||^2,
%\end{equation}

\section{Region-filtering Correlation Tracking}
\label{sec:blind}

%We introduce a spatial map to filter training samples within the standard CF formulation. Derived from the objective function, a correlation filter can be computed in the Fourier domain.
In this section, we give a detailed description of our proposed method. Firstly, we introduce the problem formulation. Secondly, we derive the optimization algorithm. Thirdly, we analyse the model update strategy. Fourthly, we summarize the tracking framework.
\subsection{Problem Formulation}
Focusing on the challenge presented above, we exploit a spatial map to filter training samples, visualized in Fig.\ref{Fig:schematic}. A spatial map is based on a priori information about the object location in the base training sample. And the spatial map can be any type, it can be a binary map eliminating the influence of the IRs, or a map similar to Gaussian penalizing the IRs, and so on. The spatial map $c$ with the size $M\times N$ is embedded into the standard CF formulation Eq.\ref{dcf_1},
\begin{equation}
\label{eq:ddcf_1}
\mathop{\mathrm{min}}_{w_l} \sum\limits_{l=1}^{d}||X_l^\top diag(c) w_l-y||_2^2+\lambda||w_l||_2^2,
\end{equation}
where $diag(c)$ is the diagonal matrix with the elements of the vector $c$ in its diagonal.
 %Examples of the spatial map $c$ are visualized in Figure \ref{Fig:schematic}.
To solve Eq.\ref{eq:ddcf_1}, we involve the introduction of an auxiliary variable $t$. In this case, Eq.\ref{eq:ddcf_1} can be identically reformulated as,
%To reformulate it by convolution operations, we introduce a variable $T$ into Eq.\ref{ddcf_1},
\begin{equation}
\begin{aligned}
&\mathop{\mathrm{min}}_{w_l,t_l} \sum\limits_{l=1}^{d}||X_l^\top t_l-y||_2^2+\lambda||w_l||_2^2 \\
&s.t. \quad t_l=diag(c)w_l
\end{aligned}
\label{eq:ddcf_2}
\end{equation}
% 同SRDCF等价，当C可逆的时候
Following this, it also can be expressed as
\begin{equation}
\begin{aligned}
&\mathop{\mathrm{min}}_{w_l,t_l} \sum\limits_{l=1}^{d}||x_l\ast t_l-y||^2+\lambda||w_l||^2 \\
&s.t. \quad t_l=c\odot w_l
\end{aligned}
\label{eq:ddcf_3}
\end{equation}

Note that, in the special case where $diag(c)$ is invertible, Eq.\ref{eq:ddcf_2} can be written
\begin{equation}
\begin{aligned}
\mathop{\mathrm{min}}_{t_l} \sum\limits_{l=1}^{d}||X_l^\top t_l-y||_2^2+\lambda||diag(c)^{-1}t_l||_2^2
\end{aligned}
\label{eq:ddcf_special}
\end{equation}
It is easy to see that Eq.\ref{eq:ddcf_special} is equivalent to SRDCF \cite{danelljan2015learning}.

\subsection{Online Optimization}
To solve the optimization problem Eq.\ref{eq:ddcf_3}, we apply the augmented Lagrangian \cite{boyd2011distributed} used in \cite{lukevzivc2016discriminative}. First, by applying Parseval's theorem to Eq.\ref{eq:ddcf_3}, the filter $w$ can be equivalently computed in the Fourier domain by minimizing the following loss function with the constraint,
\begin{equation}
\begin{aligned}
&\mathop{\mathrm{min}}_{\hat{w}_l,\hat{t}_l} \sum\limits_{l=1}^{d}||\hat{x}_l\odot \hat{t}_l-\hat{y}||^2+\lambda||w_l||^2 \\
&s.t. \quad \hat{t}_l=\widehat{c\odot w_l}
\end{aligned}
\label{eq:ddcf_4}
\end{equation}
Then, by introducing augmented Lagrange multipliers to incorporate the equality function into the loss function, the Lagrangian function is formulated as,
\begin{equation}
\begin{aligned}
%&L(\hat{t}_l, w_l, \hat{\zeta}) \\
%&= \sum\limits_{l=1}^{d}||diag(\hat{x})\hat{t}_l-\hat{y}||^2+ \lambda||w_l||^2+ [\hat{\zeta}^H(\hat{t}_l-\widehat{c\odot w_l})\\ &+\overline{\hat{\zeta}^H(\hat{t}_l-\widehat{c\odot w_l})}] + \mu||\hat{t}_l-\widehat{c\odot w_l}||^2,
&L(\hat{t}_l, w_l, \hat{\zeta}) \\
&= \sum\limits_{l=1}^{d}||diag(\hat{x}_l)\hat{t}_l-\hat{y}||^2+ \lambda||w_l||^2+ [\hat{\zeta_l}^\top(\hat{t}_l-\widehat{c\odot w_l})\\
&+\overline{\hat{\zeta_l}^\top(\hat{t}_l-\widehat{c\odot w_l})}] + \mu_l||\hat{t}_l-\widehat{c\odot w_l}||^2,
\end{aligned}
\label{eq:ddcf_5}
\end{equation}
% 是只有一个u,还是多个u? structural CF 中是多个
where $\hat{\zeta}_l$ is the Fourier transformation of the Lagrange multipliers and $\mu_l$ denotes penalty parameter which controls the rate of convergence. The optimization problem Eq.\ref{eq:ddcf_1} is transformed to minimize the Lagrangian function with the variables $\hat{t}_l, w_l $ and the Lagrange multipliers $\hat{\zeta}_l,\mu_l$. It can be optimized by iteratively solving some subproblems with closed form solutions. When a subproblem is solved for a variable, the other variables with their recent values are fixed. Thus Eq.\ref{eq:ddcf_5} can be solved by sequentially iterating the following three steps.
\vspace{2mm}
~\\
\textbf{Updating} $\bm{\hat{t}_l}$\textbf{:} Given the others, the minimization problem Eq.\ref{eq:ddcf_5} for $\left\{ \hat{t}_l\right\}_{l=1}^d$ can be decomposed into $d$ independent subproblems. The $l$-th subproblem is solved as,
%$\hat{t}$ is solved from the subproblem,
\begin{equation}
\begin{aligned}
\hat{t_l} &= \mathop{\mathrm{argmin}}_{\hat{t}_l} ||diag(\hat{x}_l)\hat{t}_l-\hat{y}||^2+ [\hat{\zeta_l}^\top(\hat{t}_l-\widehat{c\odot w_l}) \\
&+\overline{\hat{\zeta_l}^\top(\hat{t}_l-\widehat{c\odot w_l})}]+ \mu_l||\hat{t}_l-\widehat{c\odot w_l}||^2 \\
            & = \frac{\bar{\hat{x}}_l \odot \hat{y}-\hat{\zeta}_l+\mu_l\widehat{c\odot w_l}}{\bar{\hat{x}}_l \odot \hat{x}_l+\mu_l}
\end{aligned}
\label{eq:ddcf_t}
\end{equation}
where $\div$ denotes the element-wise division and $\bar{\hat{x}}_l$ indicates the complex-conjugate of $\hat{x}_l$.
% &= \frac{diag(\hat{x})^H\hat{y}-\hat{\zeta}+\mu\widehat{c\odot w}}{diag(\hat{x})^Hdiag(\hat{x})+\mu}\\
\vspace{2mm}
~\\
\textbf{Updating}  $\bm{w_l}$\textbf{:} Given the others, the minimization problem Eq.\ref{eq:ddcf_5} for $\left\{w_l\right\}_{l=1}^d$ can be decomposed into $d$ independent subproblems. The $l$-th subproblem is solved as,
%$w$ is solved from the subproblem,
\begin{equation}
\begin{aligned}
w_l &= \mathop{\mathrm{argmin}}_{w_l}  \lambda||w_l||^2 + [\hat{\zeta_l}^\top(\hat{t}_l-\widehat{c\odot w_l}) \\
&+\overline{\hat{\zeta_l}^\top(\hat{t}_l-\widehat{c\odot w_l})}]+ \mu_l||\hat{t}_l-\widehat{c\odot w_l}||^2\\
            &= \frac{c\odot \mathcal{F}^{-1}(\hat{\zeta_l}+\mu_l\hat{t_l})}{\lambda +\mu_l c \odot c}
\end{aligned}
\label{eq:ddcf_w}
\end{equation}
where $\div$ denotes the element-wise division.
\vspace{2mm}
~\\
\textbf{Updating Multiplier}  $\bm{\hat{\zeta}_l}$\textbf{:} We update the Lagrange multipliers as,
\begin{equation}
\begin{aligned}
\hat{\zeta}_l\leftarrow \hat{\zeta}_l + \mu_l (\hat{t}_l - \widehat{c \odot w_l})
\end{aligned}
\label{eq:ddcf_I}
\end{equation}
\begin{equation}
\mu_l = \mathrm{min}(\mu_{max},\beta\mu_l)
\label{eq:ddcf_mu}
\end{equation}
In experiment, we set $\mu_1 = \cdot\cdot\cdot=\mu_l=\cdot\cdot\cdot=\mu_d = \mu$.

%The solution shows that the spatial information of the training samples can be maintained. And
The full derivation is presented in the supplementary material.
\subsection{Model Update Strategy}
%Similar to \cite{henriques2015high}, we update the  filter $w_l$ by a simple linear interpolation. But we more deeply investigate the linear interpolation during the tracking.
In visual tracking, the linear interpolation is widely used. Here, we more deeply investigate the linear interpolation during the tracking.
%\begin{equation}
%w^t=(1.01-\eta)w^{t-1} + \eta w
%\end{equation}

The filter $w^k$ used for the next $k+1$ frame can be reformulated as
\begin{equation}
\begin{aligned}
w^k=&(\varrho -\alpha)^{k-1}w^1_{*}+(\varrho -\alpha)^{k-2}\alpha w^2_{*} \\
&+...+(\varrho -\alpha)\alpha w^{k-1}_{*}+\alpha w^k_{*}
\end{aligned}
\end{equation}
%where $w^k_{*}$ is the filter which is obtained from the $k$th frame and $\varrho $ is the sum of the update rates every frame. When $k$ is less than $log_{\varrho -\alpha}\alpha$, the weight of $w^1_{*}$ is greater than the weight of $w^k_{*}$. The $w^1_{*}$ is absolutely accurate, and inaccurate tracking result easily causes a deterministic failure when the target is severely occluded. Therefore, to strengthen the robustness, we promote the weight proportion of the $w^1_{*}$ by setting $\varrho=1.01$, when the learning rate $\alpha$ is conventionally set to 0.02.
where $w^k_{*}$ is the filter which is obtained from the $k$th frame and $\varrho $ is the sum of the update rates every frame. When $k$ is less than $log_{\varrho -\alpha}\alpha$, the weight of $w^1_{*}$ is greater than the weight of $w^k_{*}$. An inaccurate tracking results easily causes a deterministic failure when the target is severely occluded, in the beginning of tracking. Only, $w^1_{*}$ is absolutely accurate. To strengthen the robustness, we increase the weight proportion of the $w^1_{*}$ by making $\varrho$ greater. But note that keep $\varrho-\alpha<1$. For example, when the learning rate $\alpha$ is conventionally set to 0.02, set $\varrho$ to 1.01.

 %The $w^1_{*}$ is absolutely accurate, and inaccurate tracking result easily causes a deterministic failure when the target is severely occluded. Therefore, to strengthen the robustness, we promote the weight proportion of the $w^1_{*}$ by setting $\varrho=1.01$, when the learning rate $\alpha$ is conventionally set to 0.02.

\subsection{Our Tracking Framework}
The tracking for an image sequence is mainly as follows including training, detecting.
\vspace{2mm}
~\\
\textbf{Training:} For the standard CF, there are IRs in training samples from cyclic shifts of a base training sample. Therefore, we introduce a spatial map to filter the training samples within the standard CF formulation. New training samples are needed to be extracted and a new model should be trained for predicting in the next frame, when the object is located in the current frame of the image sequence. According to Eq.\ref{eq:ddcf_1}, we train a new model by a new base training sample $x_t$. The new base training sample is centered at the object location, and its scope is approximately four times as big as the target's. Here, $t$ is the number of the current frame.
% 初始化是不是应该交代清楚
\vspace{2mm}
~\\
\textbf{Detecting:} In the detection stage, the tracker estimates the target location as a new frame comes. Centered at the target location in previous frame, a base detecting sample $z$ is extracted from the new frame. For the base detecting sample $z$, we compute the response $S$ as,
\begin{equation}
\label{ddcf_detect}
S(z) = \mathcal{F}^{-1}\left(\sum\limits_{l=1}^{d}\hat{z}_l\odot \widehat{w_l\odot c}\right),
\end{equation}

The location with the maximum in the response is about the translation. And we apply the sub-grid interpolation strategy \cite{danelljan2015learning} to maximize the scores of the response. Meanwhile, multiple base training samples with different sizes centered around the target are used for estimating the target scale. Let $M \times N$ denote the fiducial search size and $\kappa$ be the current change factor of the target. The size of a base training sample ${x_{a^r}}$ is $M\kappa a^r \times N\kappa a^r$. Here, $a$ is the scale increment factor, $r\in \left\{{\left \lfloor \frac{1-s}{2} \right \rfloor,...,\left \lfloor \frac{s-1}{2} \right \rfloor}\right\}$ and $s$ denotes the number of scales. A base training sample ${x_{a^r}}$ is resized to the fiducial search size before the model computation.
The current change factor of the target $\kappa$ is updated according to scale change, as the target is located in the frame.
% 以上帧的目标中心位置为中心采样,
% 写详细
%\vspace{2mm}
%~\\
%\textbf{Model Updating:}

An overview of our tracker is summarized in Algorithm 1.

\begin{algorithm}[htb]
  \caption{RFCT algorithm}
   \label{alg:AlgTrakcer}
    \begin{algorithmic}[1]
      \Require
        Frame sequence $\left\{I\right\}_{k=1}^K$, the initial location $b_1$, desired response $y$, spatial map $c$
      \Ensure
        Tracking result for each frame $b_k$
      \Repeat
        \State Crop several search regions with different scales from the frame $I_k$, which are centered at the last location $b_{k-1}$, and extract features of each region $z_r$.
        \State Calculate the response of each base sample $z_r$ using Eq.\ref{ddcf_detect} to estimate the location and scale of the target.
      \While {stop condition}
        \State Update $\hat{t}_l$ using Eq.\ref{eq:ddcf_t}
        \State Update $w_l$ using Eq.\ref{eq:ddcf_w}
        \State Update $\hat{\zeta}_l$, $\mu_l$ using Eq.\ref{eq:ddcf_I} and Eq.\ref{eq:ddcf_mu}
      \EndWhile
      \State Update the filter by the linear interpolation.
      \Until the end of the sequence
    \end{algorithmic}
\end{algorithm}
\section{Experiment Results}
%In this section, we compare the proposed method with trackers based on conventional features and trackers based on deep features respectively. Results are reported on OTB-2013 and OTB-2015.
In this section, we first present the experimental setup. Then, we analyse the effects of three spatial maps and the increasing of the weights for accurate filters. Finally, we compare the proposed method with trackers based on conventional features and trackers based on deep features respectively.
\subsection{Experimental Setup}
Here, we present experimental setup, datasets and evaluation metrics.
%Then, we compare the proposed method with state-of-the-art trackers in quantitative and qualitative ways.
\vspace{2mm}
~\\
\textbf{Parameters Setup:}
%The spatial map can be freely set, based on prior information. It can be a binary mask map, or a Gaussian map and so on.
%In our experiment, we alter the map with the zero values beyond 1.6 times the target region, based on a quadratic map $c(p,q) = [\nu +\delta (p/W)^2+\delta (q/H)^2]^{-1}$.Here, $W\times H$ is the target size. Meanwhile, $\nu$ is a parameter which controls the maximum value of the map $c$, and $\delta$ is a penalty factor. In my experiment, $\nu$ is set to 0.1.
Similar to \cite{danelljan2016eco,lukevzivc2016discriminative,danelljan2016beyond}, we apply HOG and Color-Names features multiplied by a Hann window for the filter $w$. A cell size of 4 $\times$ 4 pixels is employed for HOG features, and the image region area of a base training sample is $4^2$ times the target size. The label $y$ is a Gaussian map with a standard deviation proportional to the target size.The sum of update rates is set to 1.01, and the learning rate is set to 0.02. The augmented Lagrangian optimization parameters are set to $\mu^{(0)}=5,\beta = 3$ and $\mu_{max} = 20$. In addition, we set the number of iterations is 8.
%All parameters are fixed for all datasets. To reduce the effect of training sample containing periodic repetitions, we employ a base training sample multiplied by a Hann window \cite{bolme2010visual}.
%\cite{danelljan2015learning}
% 为什么选择quadratic function？
\vspace{2mm}
~\\
\textbf{Datasets:} We evaluate the proposed tracker on two datasets: OTB-2013 \cite{wu2013online} and OTB-2015 \cite{wu2015object}. There are 50 videos annotated with ground truth bounding boxes and various visual attributes in OTB-2013. The recently introduced OTB-2015 is an expansion of OTB-2013, containing 100 sequences.
% 11 个属性
\vspace{2mm}
~\\
\textbf{Evaluation Metrics:} The proposed method is compared with state-of-the-art trackers, employing the evaluation metrics and code provided by the benchmark datasets. Our approach is quantitatively evaluated in two aspects in \cite{wu2013online}, precision rate and success rate. Distance precise (DP) is widely used for precision rate, the percentage of frames whose distance between the center locations of the tracked targets and the manually labeled ground truths is within a given threshold distance of the ground truth. The threshold is commonly set to 20 pixels. Overlap precise (OP) is applied for success rate, the ratios of successful frames whose overlap between the tracked bounding box and the ground truth bounding box is larger than a given threshold. The threshold is varied from 0 to 1. The specific threshold 0.5 and the area under curve (AUC) of a success plot are usually used to rank the tracking algorithms respectively.

\subsection{Analysis of RFCT}
The values of the spatial map are not restricted in our proposed method. Therefore, we demonstrate the effects of three spatial maps by experiments on OTB-2013. When none of the values in the spatial map is zero, our proposed Eq.\ref{eq:ddcf_special} is equivalent to SRDCF \cite{danelljan2015learning}. Therefore, the quadratic function in SRDCF is chose to design a spatial map $c(p,q) = [\nu +\delta (p/W)^2+\delta (q/H)^2]^{-1}$, which is named Rquadratic Map. Here, $W\times H$ is the target size.
We aims to eliminate the interference of the IR. At the same time, considering severe deformation, the surrounding nearly around the target may bring the discriminative information. We alter the above Rquadratic Map, setting the values beyond the region with 1.6 times the target region to zero. The altered map is named Our Map. In addition, we take a binary map with 1 in the target region and 0 in the other region for comparison, named Binary Map. The tracking results based on the above three maps are summarized in Table \ref{tab:analysis}.

As the shown in Table \ref{tab:analysis}, Our Map is a little better than the other two maps. Further, based on Our Map, we implement a tracker named RFCT increasing the weight proportions of accurate filters. In Table \ref{tab:analysis}, the result clearly shows that the increase brings the positive effect.

\begin{table}[htbp]
\center
\vspace{2mm}\caption{Analysis of our approach on OTB-2013. We report the area-under-the-curve (AUC) scores (\%). The \textcolor[rgb]{1,0,0}{first} and \textcolor[rgb]{0,1,0}{second} rank values are highlighted in color.}
%\small
\newcommand{\tabincell}[2]{\begin{tabular}{@{}#1@{}}#2\end{tabular}}
\begin{tabular}{p{2cm}<{\centering}|p{2cm}<{\centering}|p{3cm}<{\centering}|p{2cm}<{\centering}|p{1.8cm}<{\centering}}
 \hline
 \multirow{2}{*}{} &
 \multicolumn{3}{c|}{\tabincell{c}{Our trackers based on different maps}} &
 \multicolumn{1}{c}{\multirow{2}{*}{RFCT}} \\
 \cline{2-4}
   & Binary Map & Rquadratic Map & Our Map & \multicolumn{1}{c}{} \\
 \hline
 OTB-2013 & 58.0 & 61.5 & \textcolor[rgb]{0,1,0}{62.6}   & \textcolor[rgb]{1,0,0}{65.9} \\
 \hline
 \end{tabular}
 \label{tab:analysis}
\end{table}

%\begin{table}[htbp]
%\small
%%\renewcommand{\raggedright}{\leftskip=0pt \rightskip=0pt plus 0cm}
%%\newcommand{\tabincell}[2]{\begin{tabular}{@{}#1@{}}#2\end{tabular}}
% \centering
%  \begin{tabular*}{|c|c|c|c|}
%  \hline
%    \multirow{2}{1cm}{}& \multicolumn{3}{|c|}{trackers based on different maps}& \multirow{2}{lcm}{\textbf{RFCT}} & Binary Map &Quadratic Map &Our Map \\
%
%  \hline
%    OTB-2013& \textcolor[rgb]{0,1,0}{58.0}       & \textcolor[rgb]{0,0,1}{61.5} & 62.6       & \textcolor[rgb]{1,0,0}{65.9}  \\
%  \hline
% \end{tabular*}
% \vspace{0.5mm}\caption{Success rate of RFCT compared to the above trackers based on deep features at an overlap threshold 0.5. The \textcolor[rgb]{1,0,0}{first}, \textcolor[rgb]{0,1,0}{second} and \textcolor[rgb]{0,0,1}{third} rank values are highlighted in color.}
% \label{tab:analysis}
%\end{table}

\subsection{Comparison with Trackers based on Conventional Features}
% ECO-HC, SRDCF, CSR-DCF, SCF, Staple$_{CA}$%为了清晰，只显示top10% 几个图，几个表？0.5的表，AUC 的%图
% 对比AUC和0.5
We evaluate the proposed method on the benchmark OTB-2013 and OTB-2015 with comparisons to 42 trackers using conventional hand-crafted features, including 41 trackers in \cite{wu2013online}, and these state-of-the-art trackers ECO-HC \cite{danelljan2016eco}, BACF \cite{kiani2017learning}, SRDCFdecon \cite{danelljan2016adaptive}, CSR-DCF \cite{lukevzivc2016discriminative},STAPLE$_{\rm{CA}}$ \cite{mueller2017context}, SRDCF \cite{danelljan2015learning}, CFBL \cite{kiani2015correlation}, LCT \cite{ma2015long}, SAMF \cite{li2014scale}, MEEM \cite{zhang2014meem}, DSST \cite{danelljan2014accurate} and KCF \cite{henriques2015high}.
% DLSSVM \cite{}, MUSTer \cite{hong2015multi}
%Staple \cite{bertinetto2016staple},

\begin{table*}[htbp]
\vspace{0.5mm}\caption{Success rates (\%) of RFCT compared with these conventional features based trackers mentioned above at an overlap threshold 0.5. For clarity, only the top 10 trackers are displayed. The top 3 rank values are highlighted by \textcolor[rgb]{1,0,0}{red}, \textcolor[rgb]{0,1,0}{green} and \textcolor[rgb]{0,0,1}{blue} respectively.}
\scriptsize
\newcommand{\tabincell}[2]{\begin{tabular}{@{}#1@{}}#2\end{tabular}}
 \centering
 %\caption{\label{tab:test}The overlap rate.}
  %\begin{tabular}{|c|c|c|c|c|c|c|c|c|c|c|}
  \begin{tabular*}{\textwidth}{@{\extracolsep{\fill}} p{2cm}<{\centering}cccccccccc}
% \begin{tabular}{|p{1.7cm}<{\centering}|p{1cm}<{\centering}|p{0.8cm}<{\centering}|p{0.8cm}<{\centering}|p{0.8cm}<{\centering}|p{0.8cm}<{\centering}|p{1cm}<{\centering}|p{0.8cm}<{\centering}|p{0.8cm}<{\centering}|p{1.2cm}<{\centering}|p{0.8cm}<{\centering}|}

  %\toprule
  \hline
  %\rowcolor{mygray}
           &\textbf{RFCT}&\tabincell{c}{ECO-HC \\ \cite{danelljan2016eco}}& \tabincell{c}{SRDCFdecon \\ \cite{danelljan2016adaptive}} & \tabincell{c}{BACF \\ \cite{kiani2017learning}} & \tabincell{c}{CSR-DCF \\ \cite{lukevzivc2016discriminative}} & \tabincell{c}{STAPLE$_{\rm{CA}}$ \\ \cite{mueller2017context}} &\tabincell{c}{SRDCF \\ \cite{danelljan2015learning}}  & \tabincell{c}{LCT \\ \cite{ma2015long}}  &\tabincell{c}{SAMF \\ \cite{li2014scale}}  & \tabincell{c}{MEEM \\ \cite{zhang2014meem}}  \\
  %\midrule
  \hline
    OTB-2013& \textcolor[rgb]{1,0,0}{84.1}       & 81.0 & \textcolor[rgb]{0,0,1}{81.4}       & \textcolor[rgb]{0,1,0}{84.0}  & 75.6   & 76.5      & 78.1   & {81.3}& 73.2  & 69.6 \\
  \hline
    OTB-2015& \textcolor[rgb]{0,0,1}{77.2}       & \textcolor[rgb]{1,0,0}{78.4} & {76.6}       & \textcolor[rgb]{0,1,0}{77.6}  & 70.8   & 72.8      & 72.9   & 70.1& 67.5  & 62.2  \\
  \hline
 \end{tabular*}
 \label{tab:ConventionalCmpThreshold}
\end{table*}

Table \ref{tab:ConventionalCmpThreshold} reports the success evaluation of our method and those conventional trackers at an overlap threshold 0.5. For the sake of clarity, we present the top 10 trackers in the table. Among these trackers, ECO-HC and BACF perform better results with an overlap precise of 81.0\%, 84\% on OTB-2013 and an overlap precise of 78.4\%, 77.6\% on OTB-2015 respectively. The proposed RFCT performs well with an overlap precise of 84.1\% on OTB-2013 and an overlap precise of 77.2\% on OTB-2015. The results show that our method achieves comparable results as ECO-HC. Our RFCT performs well than other trackers, the details are as followed. (1) %Compared with these correlation filter tracking algorithms apart from the ECO-HC, the proposed DDCF method performs well against
Among these conventional correlation filter tracking algorithms apart from the BACF and ECO-HC, SRDCFdecon and SRDCF are 2 top existing trackers. On OTB-2013, SRDCFdecon and SRDCF achieve OP of 81.4\%, 78.1\% respectively. Our approach generates better tracking results by 2.7\% and 6\%. On OTB-2015, SRDCFdecon and SRDCF separately achieve OP of 76.6\%, 72.8\%. Our method performs slightly better tracking results by 0.6\% and 4.4\%. (2) Compared with these trackers except correlation filter tracking algorithms, the proposed method achieve better tracking performance against MEEM the best tracker by 14.5\% on OTB-2013 and 15.0\% on OTB-2015.
% 对比基于CFtrackers，对比其他
%从 OTB2013，OTB2015 分别对比
% 相似方法对比，分析结果好的原因
% note that， 我们的方法还可以应用别的策略。
\begin{figure}
\centering
\includegraphics[width=1.6in]{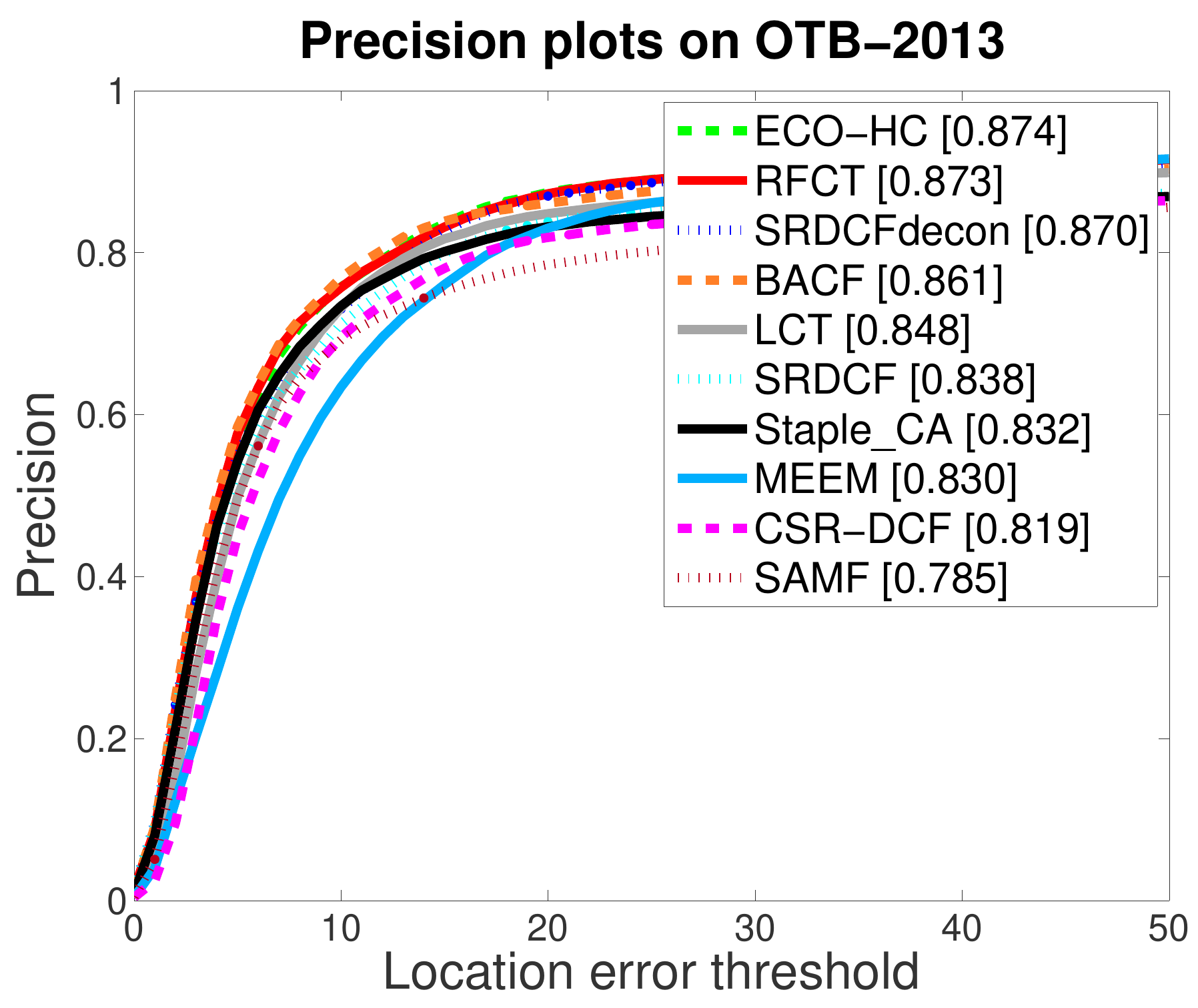}\quad
\includegraphics[width=1.6in]{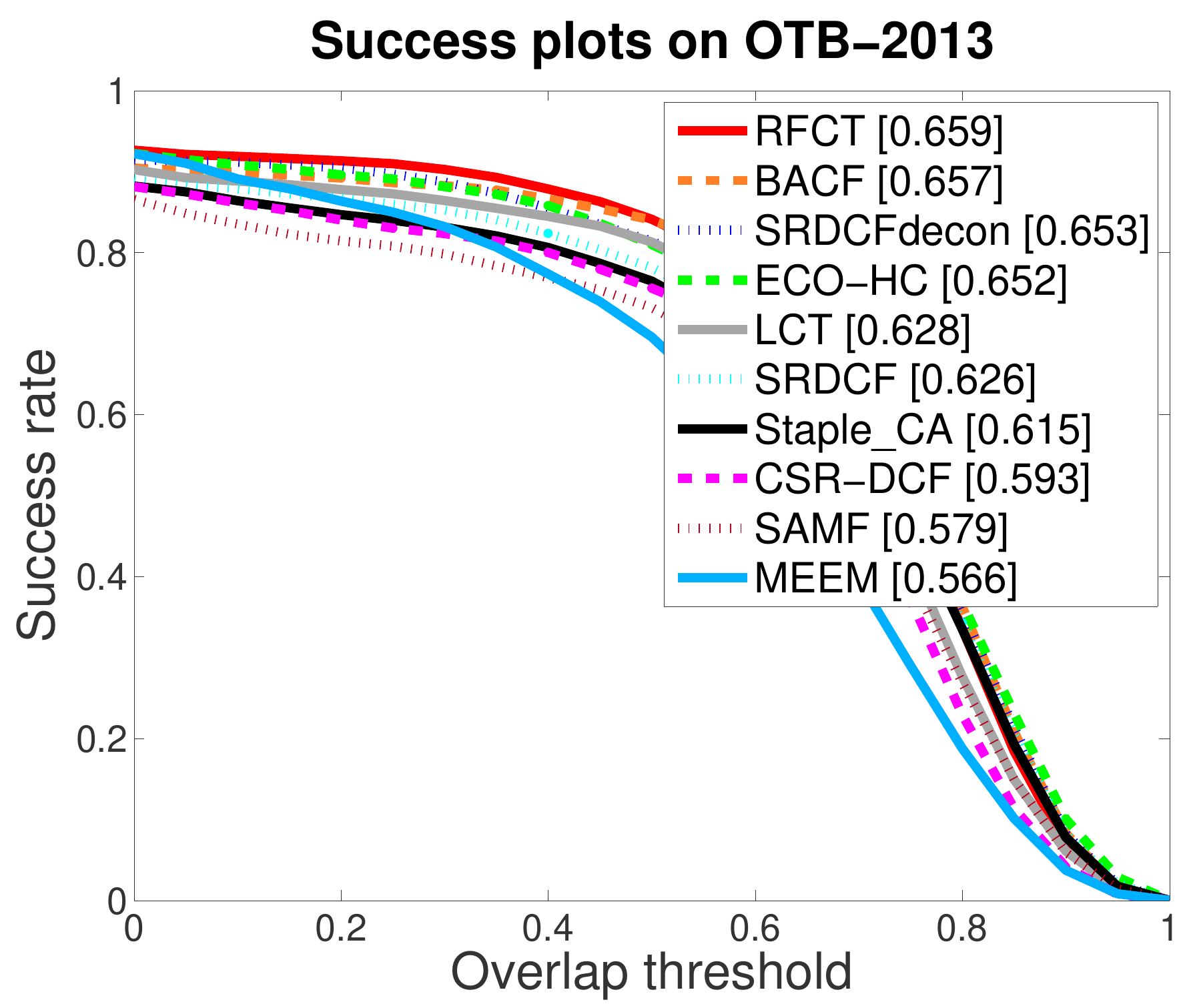}
\includegraphics[width=1.6in]{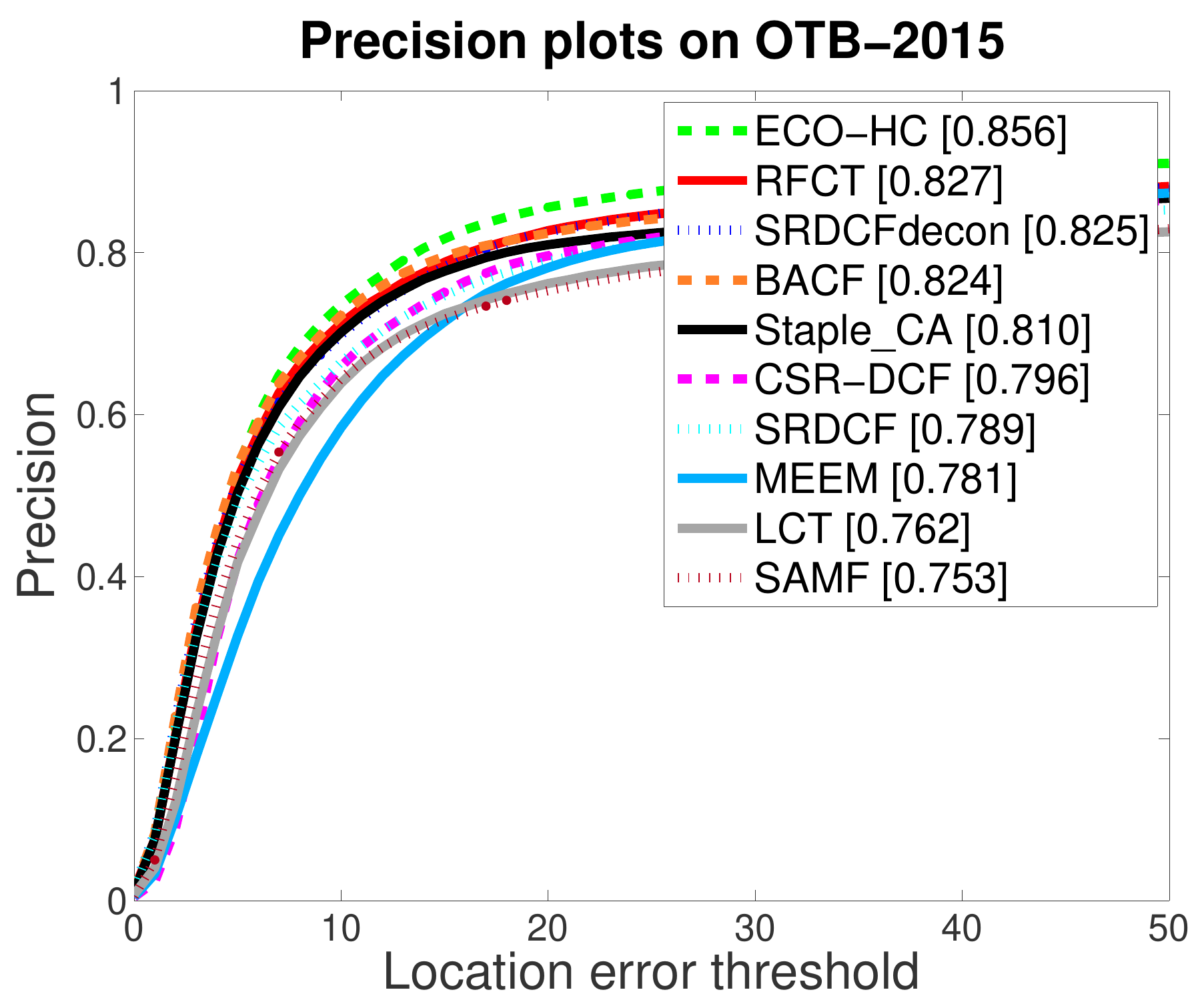}\quad
\includegraphics[width=1.6in]{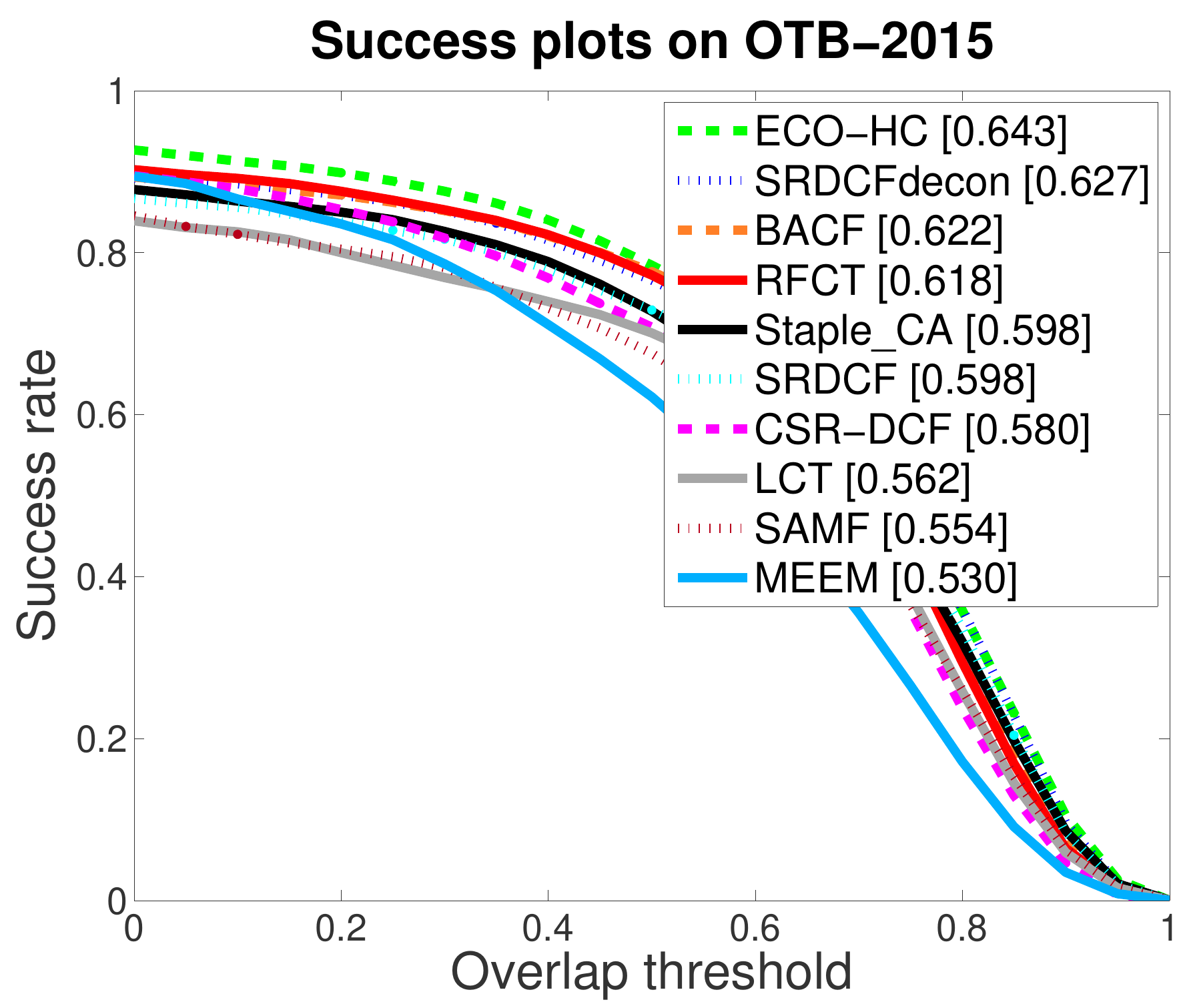}
\caption{Precision and success plots reporting a comparison with the conventional features based trackers on OTB-2013 and OTB-2015 datasets. For clarity, we only present the top 10 trackers in each plot. The area-under-the-curve (AUC) score of each tracker is reported in a bracket for success plots.
}
\label{Fig:ConventionalCmp}
\end{figure}

Fig. \ref{Fig:ConventionalCmp} compares RFCT with these conventional tracking algorithms, containing precision and success plots illustrating distance precise (DP) and overlap precise (OP) on OTB-2013 and OTB-2015. For success plots, AUCs are reported in brackets. In both precision and success plots on OTB-2013, the proposed RFCT approach achieves slightly better performance.
%achieves the distance precise rate and AUC (0.878,0.666), closely followed by ECO-HC (0.874,0.652) and SRDCFdecon (0.870,0.653).
%followed by BACF, SRDCFdecon and ECO-HC.
On OTB-2015, our tracker shows comparable results as BACF, SRDCFdecon and outperforms other trackers. %Note that, the proposed RFCT tracker can be improved by considering other model updating strategy.
 %such as that used in the ECO-HC tracker \cite{danelljan2016eco}.
 These results demonstrate the importance of filtering training samples to learn a more robust tracker. This evaluation also shows that the proposed method is effective.
%This is mainly because our approach 从训练样本的角度解决这个问题
%slightly outperforms
\begin{figure*}
\centering
\includegraphics[width=1.6in]{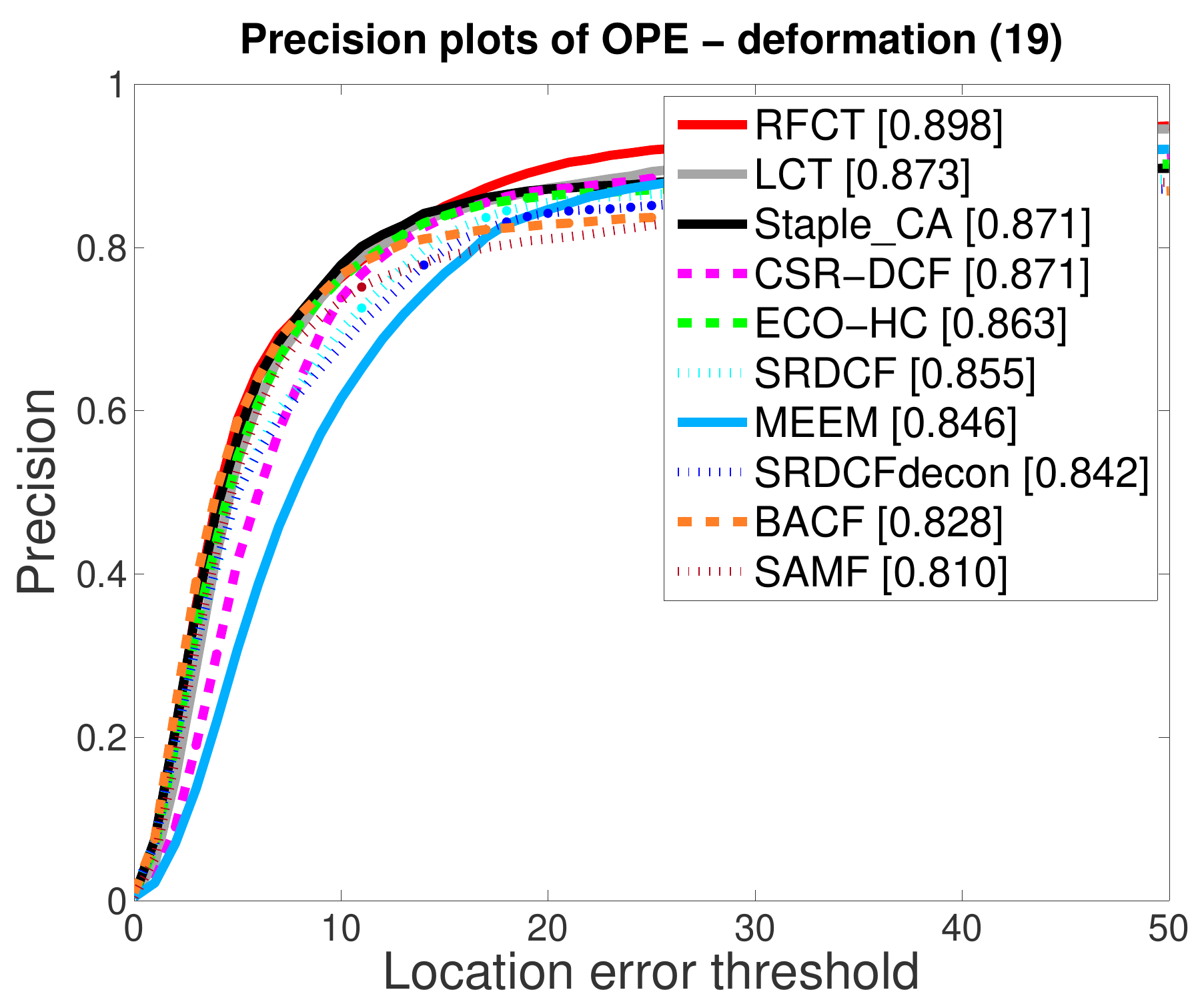}\hfill
\includegraphics[width=1.6in]{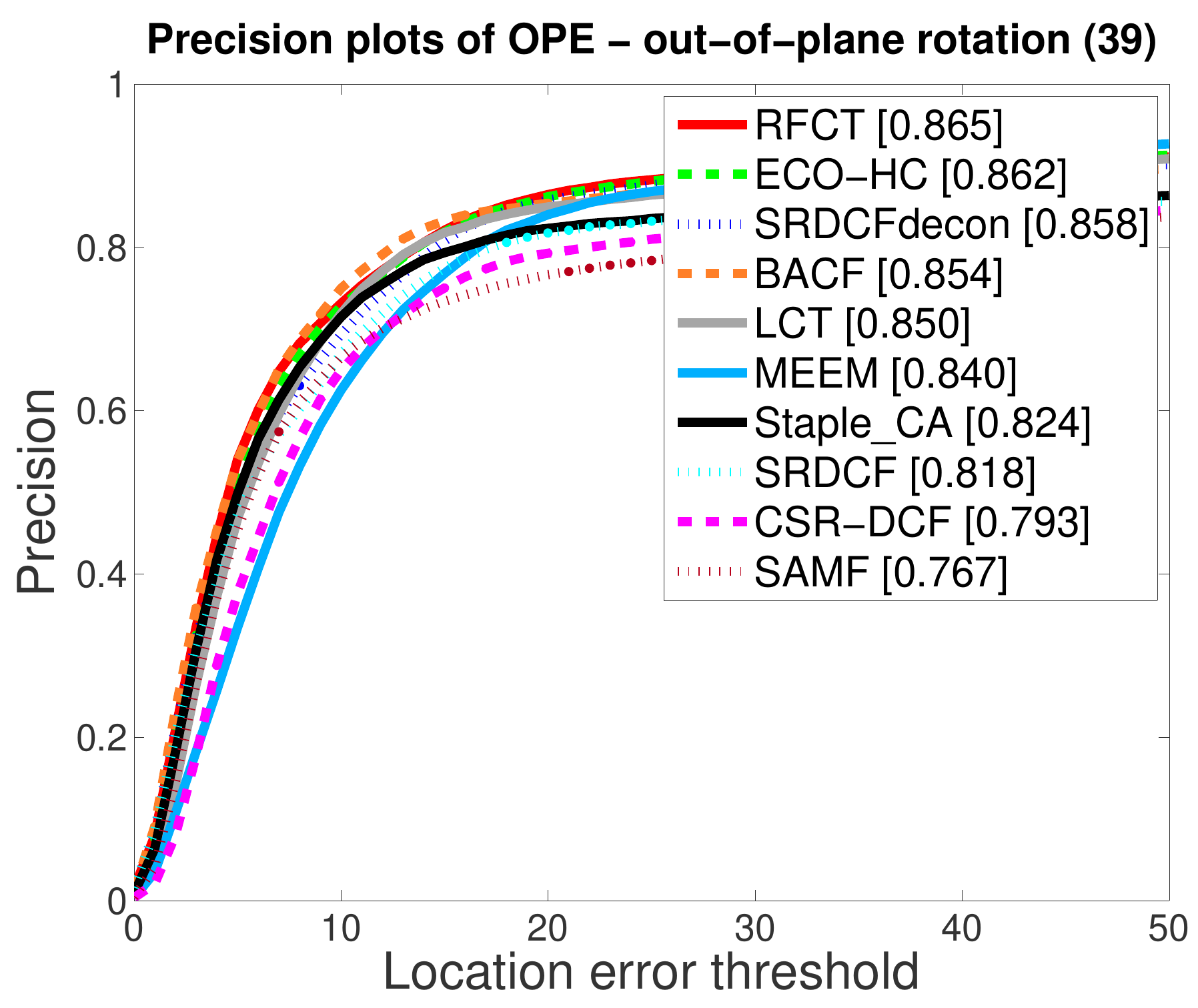}\hfill
\includegraphics[width=1.6in]{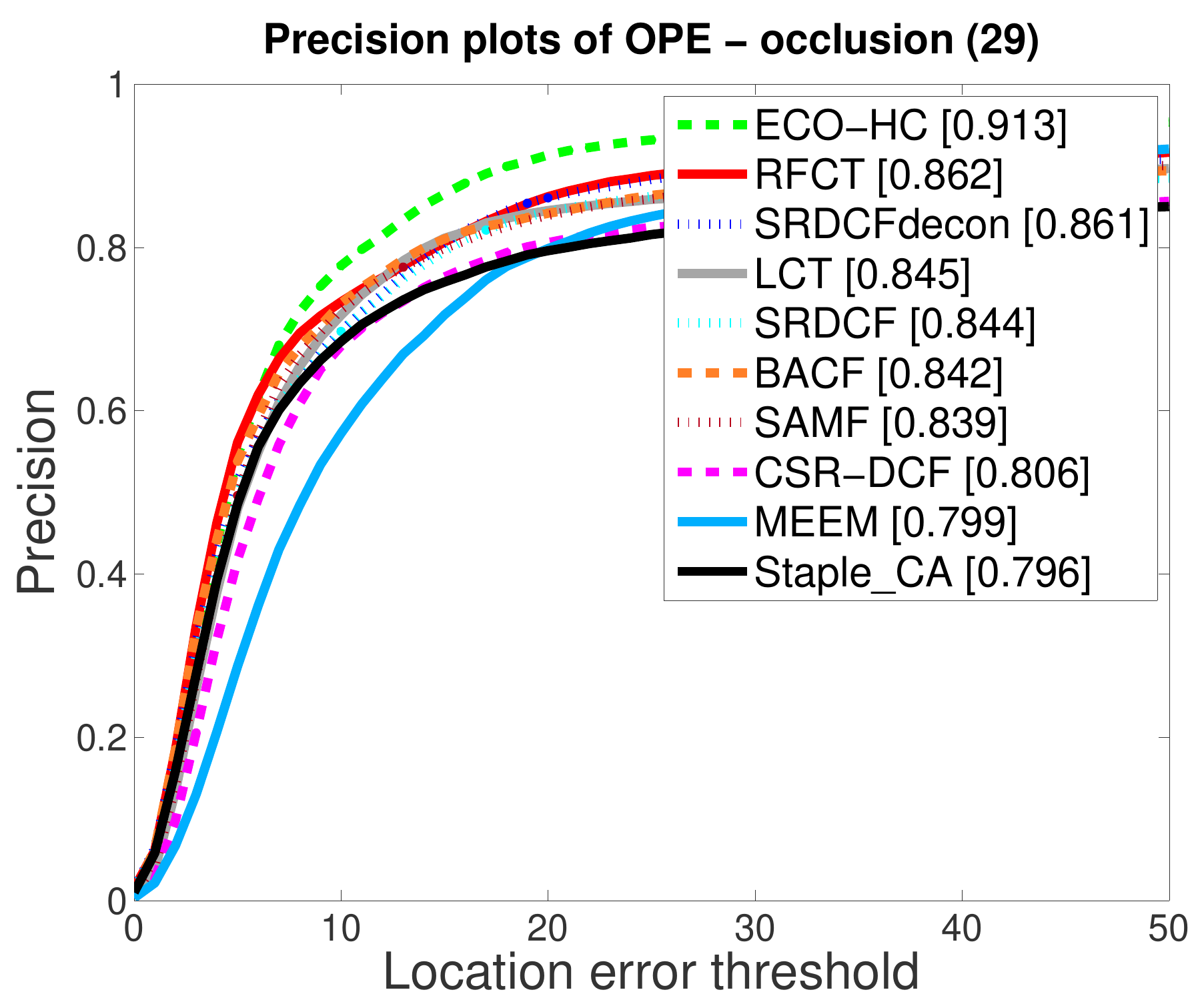}

\includegraphics[width=1.6in]{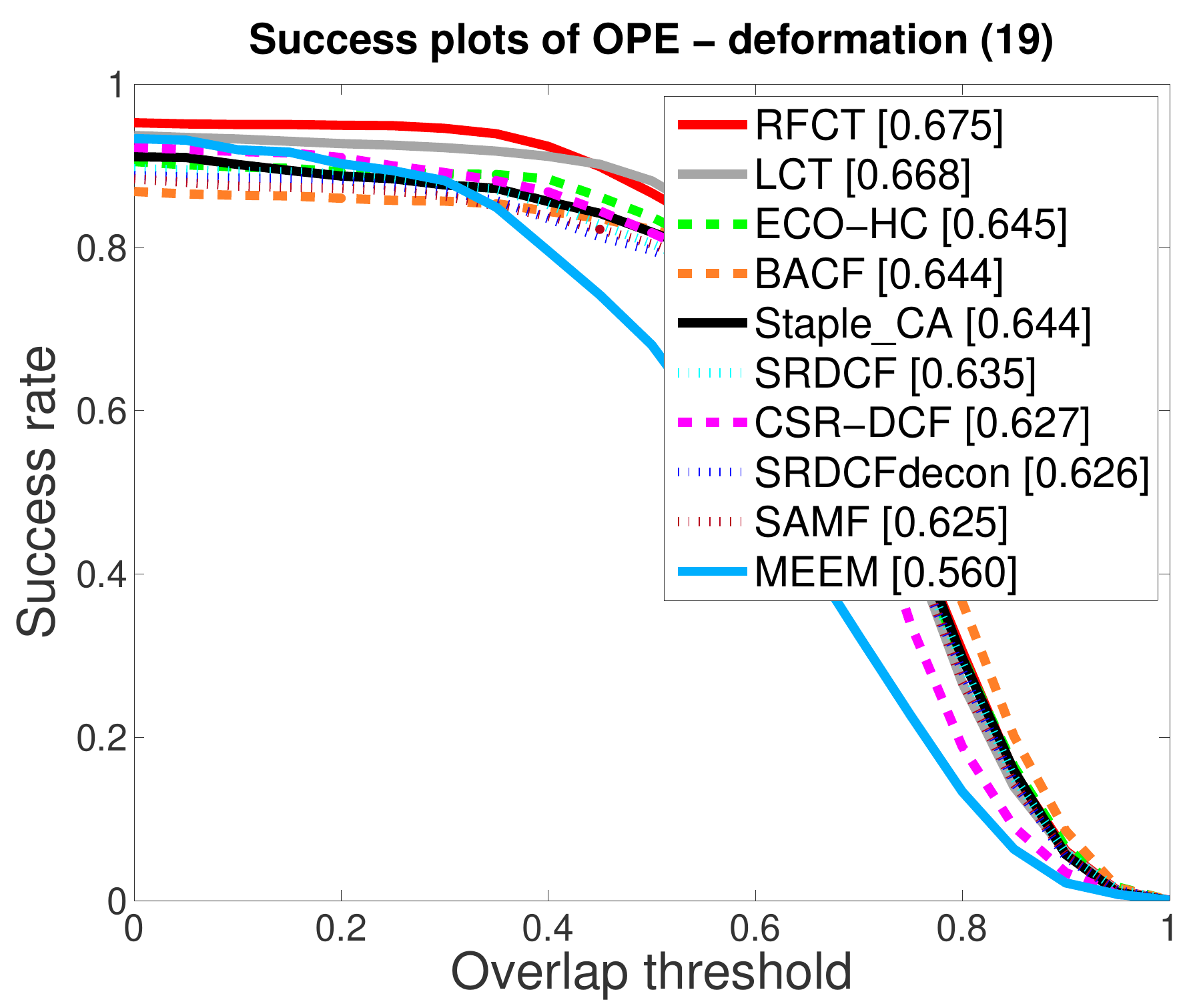}\hfill
\includegraphics[width=1.6in]{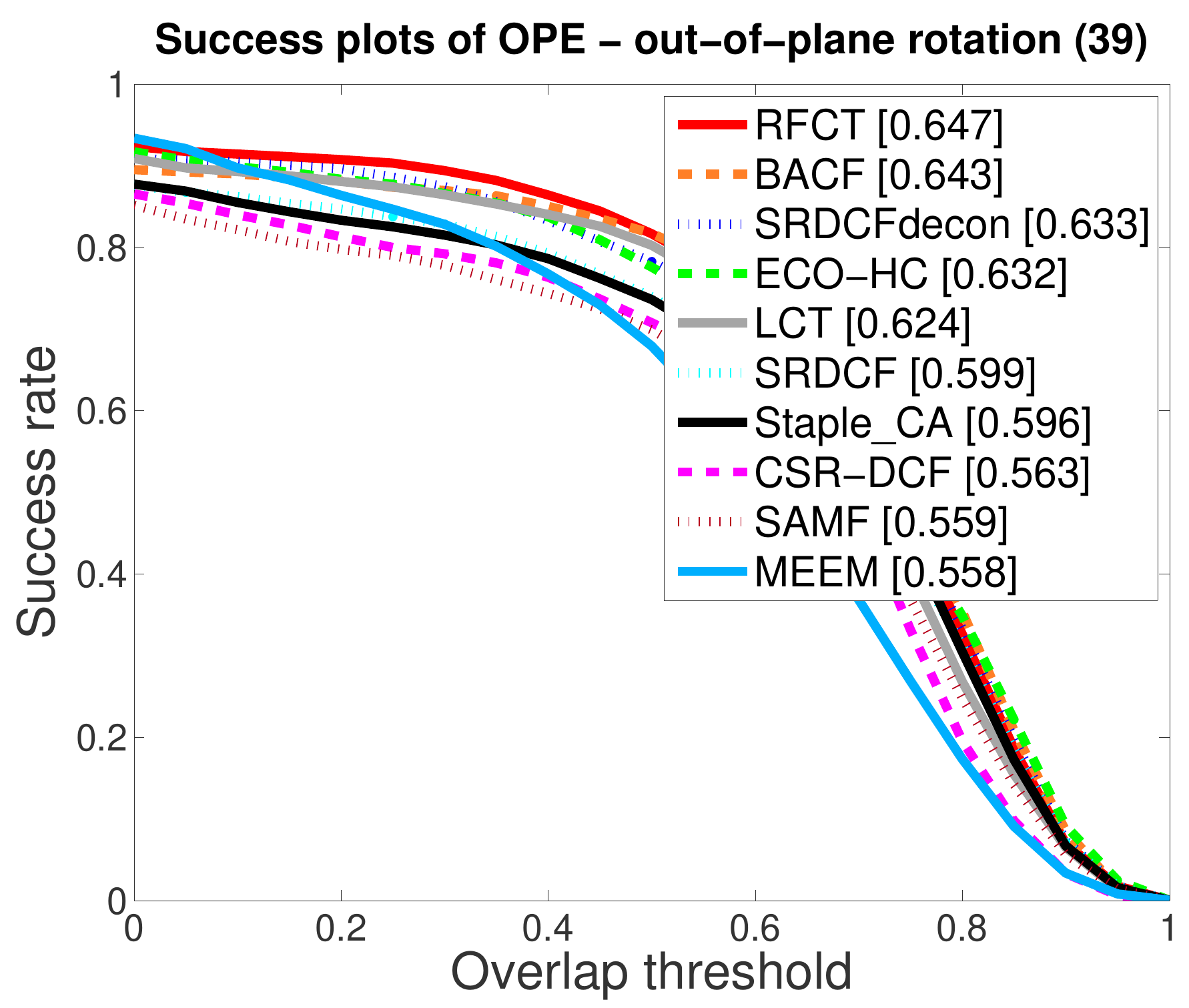}\hfill
\includegraphics[width=1.6in]{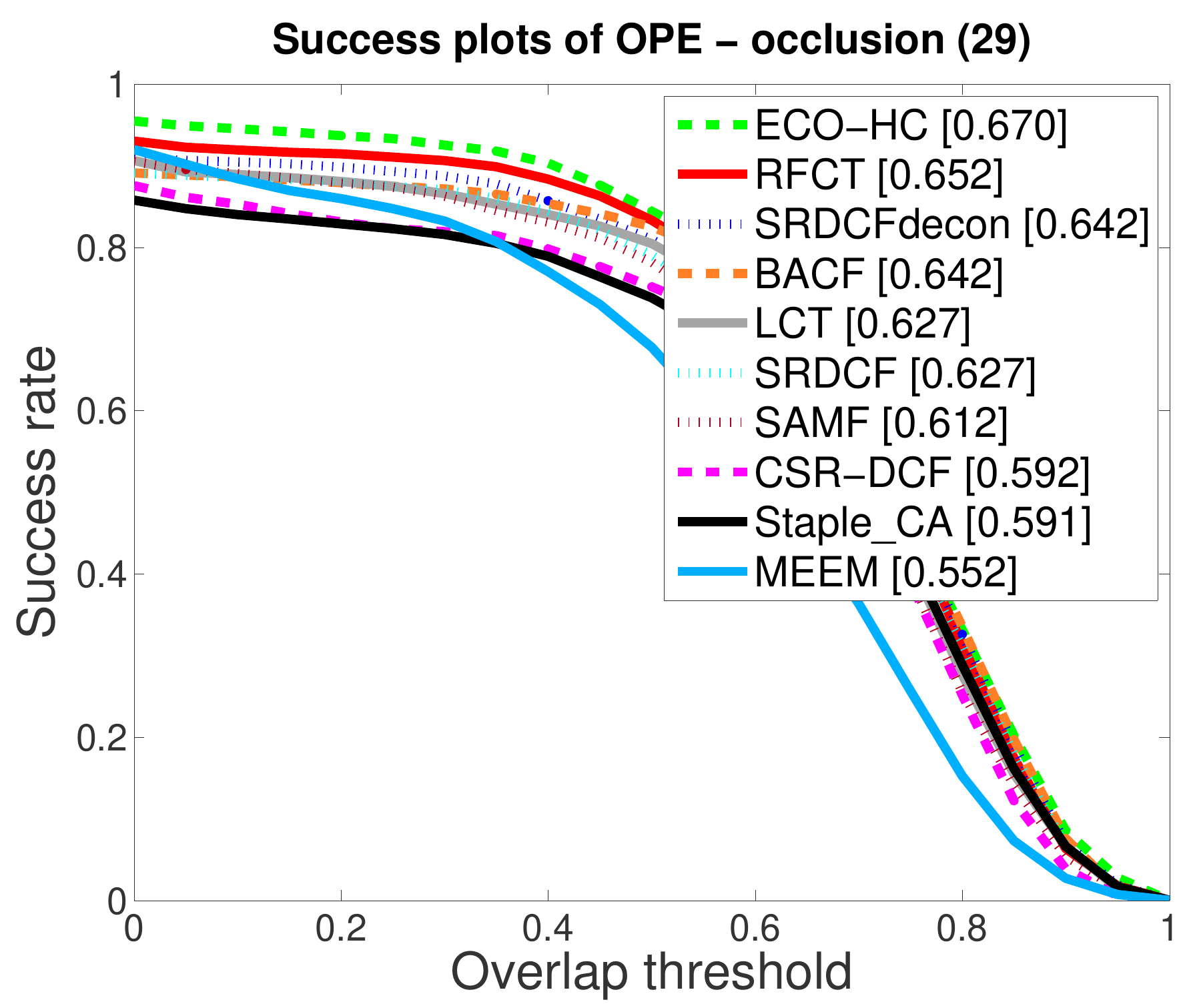}

\caption{Attribute based evaluation. Precision and success plots compare the proposed RFCT method with these conventional features based trackers over four tracking challenges on OTB-2013. AUCs are reported in brackets for success plots. For clarity, only the top 10 trackers are showed in each plot. }
\label{Fig:AttributesCmp}
\end{figure*}

\vspace{2mm}
~\\
\textbf{Attribute Based Comparison:} We show an attribute based evaluation of the proposed approach on OTB-2013. The dataset videos are annotated with 11 different attributes namely occlusion, deformation, motion blur, fast motion, in-plane rotation, out-of-plane rotation, illumination, out-of-view, variation, background clutter, and low resolution. A tracker can be analysed in the 11 different aspects. Due to space constraints, we present precision and success plots of OPE for 4 attributes in Fig. \ref{Fig:AttributesCmp} and more results can be found in the supplementary material. And for clarity, only the top 10 trackers are showed in each plot. In cases of deformation and out-of-plane rotation, the proposed algorithm performs well against other trackers. In case of occlusion
%and background clutter,
, our tracker shows better results than SRDCFdecon.
%our tracker achieve a gain of 2.4\% compared to SRDCFdecon for AUC score. Among the existing trackers, SRDCFdecon is the best trackers in case of background clutter. Our approach achieves a gain of 0.9\% over SRDCFdecon.

\begin{figure}
\centering
\includegraphics[width=1.6in]{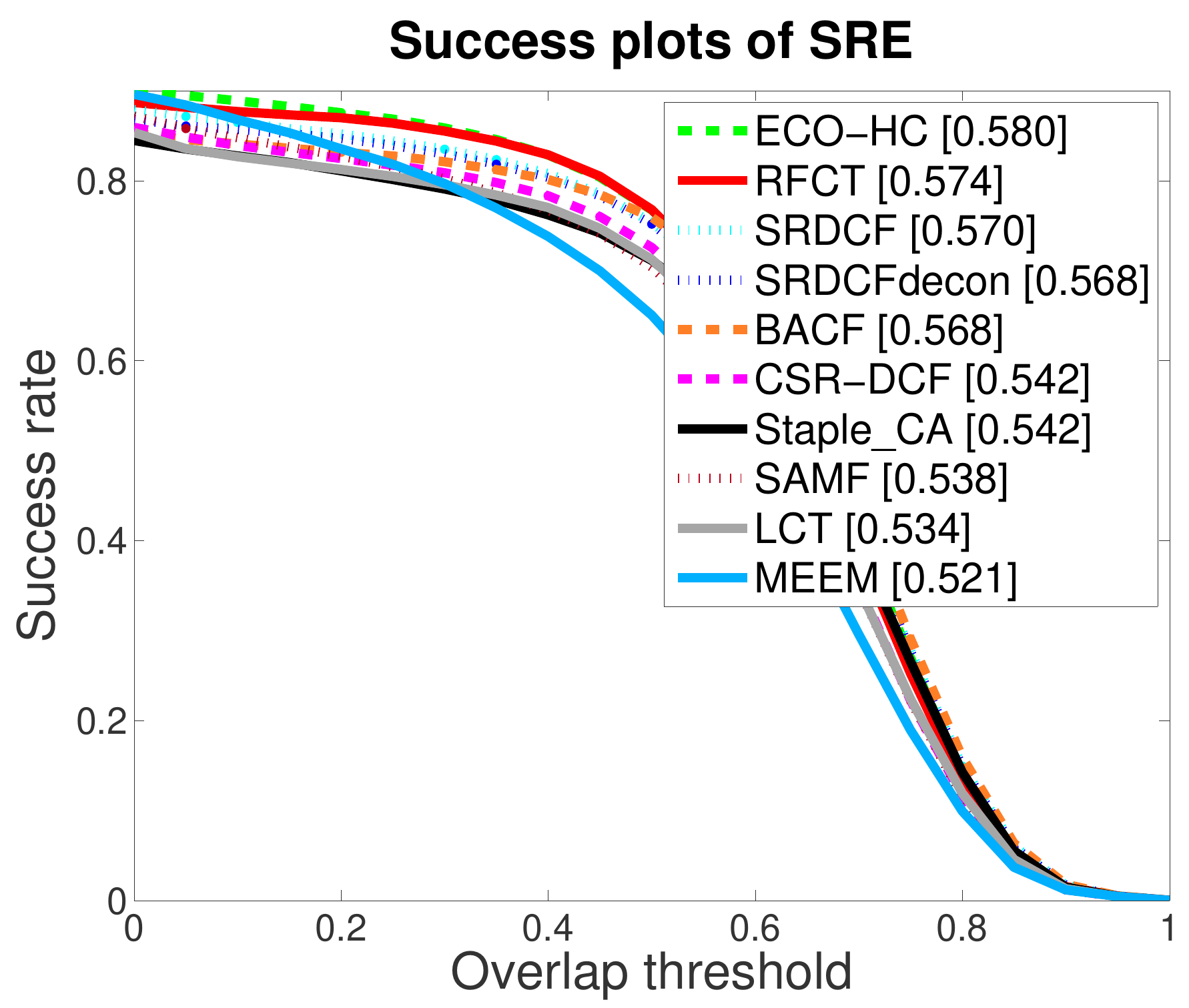}\quad
\includegraphics[width=1.6in]{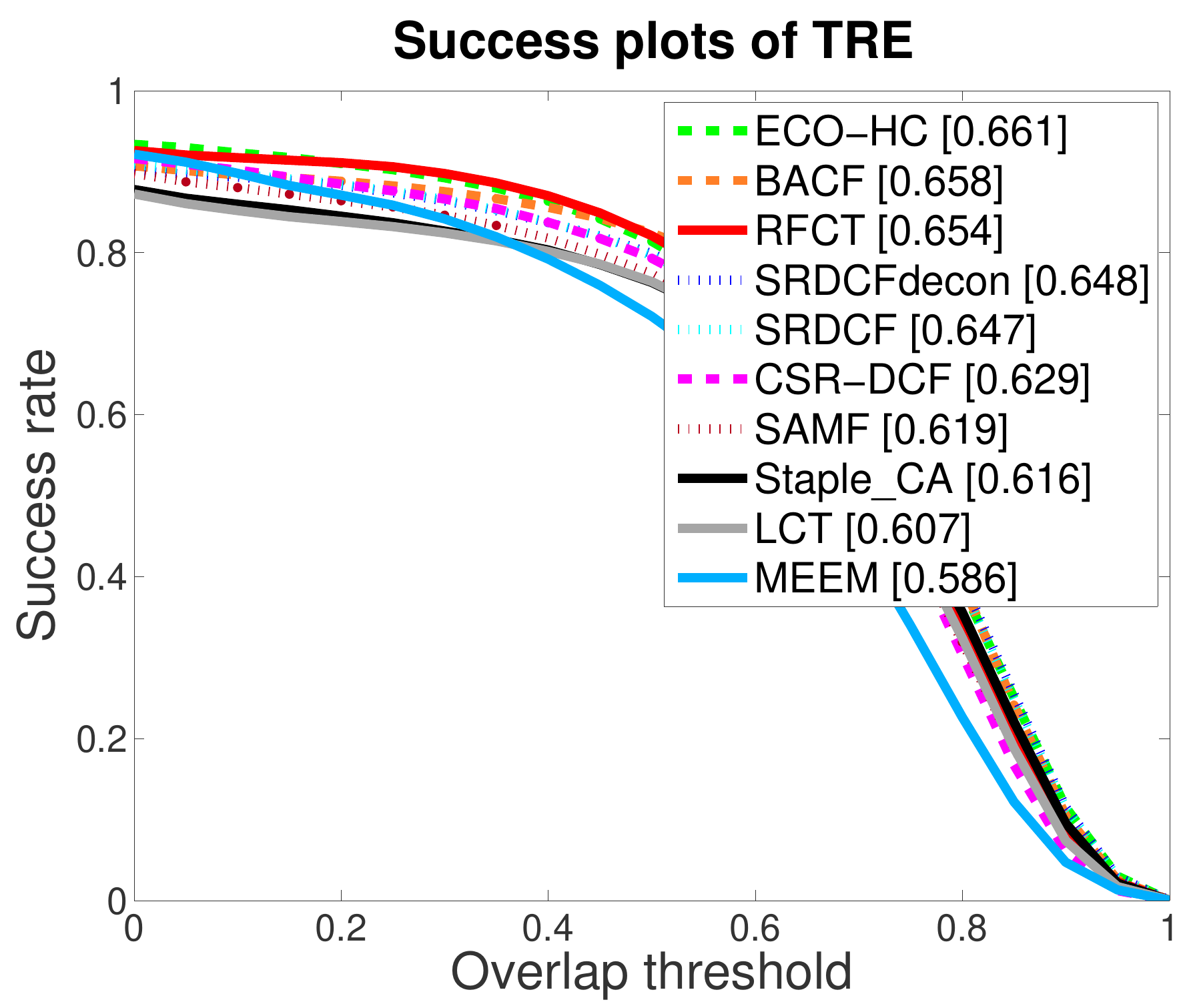}
\caption{Robustness based evaluation. The success plots on SRE and TRE compare RFCT with these above trackers based on conventional features based on OTB-2013. AUCs are reported in brackets, and only the top 10 trackers are showed in each plot for clarity.}
\label{Fig:ConventionalRobust}
\end{figure}

\vspace{2mm}
~\\
\textbf{Robustness to Initialization:} We evaluate the robustness of our approach to different temporal and spatial initialization on OTB-2013 using two metrics \cite{wu2013online}, spatial robustness evaluation (SRE) and temporal robustness evaluation (TRE). SRE evaluate the sensitivity of a tracker when initialize the tracker by different bounding boxes. TRE shows the performance of a tracker with different initializations at different start frames in a video. Fig. \ref{Fig:ConventionalRobust} shows the SRE and TRE success plots of RFCT with these conventional feature base trackers mentioned above on OTB-2013. The proposed RFCT tracker achieves comparable AUC scores as ECO-HC and BACF, and performs well against other trackers.

\subsection{Comparison with Correlation Trackers based on Deep Features }
% DeepSRDCF, HDT, MCPF(Multi-task Correlation Particle Filter for Robust Object Tracking) CCOT

%
We also compare the proposed approach based on conventional features with the several state-of-the-art trackers based deep features including MCPF \cite{zhang2017multi}, DeepSRDCF \cite{danelljan2015convolutional}, HDT \cite{qi2016hedged} on OTB-2013 and OTB-2015 dataset.

Table \ref{tab:deepCompThreshold} shows success rates of the conventional features based RFCT tracker compared with the several deep feature based trackers at an overlap threshold 0.5. Among the existing methods, MCPF performs best with the AUC scores of 85.8\% on OTB-2013 and 78.0\% on OTB-2015. Our approach achieves the AUC scores of 84.1\% on OTB-2013 and 77.2\% on OTB-2015. The proposed method shows comparable results as MCPF. And our approach slightly outperforms DeepSRDCF and HDT on the both datasets.

\begin{figure}
\centering
\includegraphics[width=1.6in]{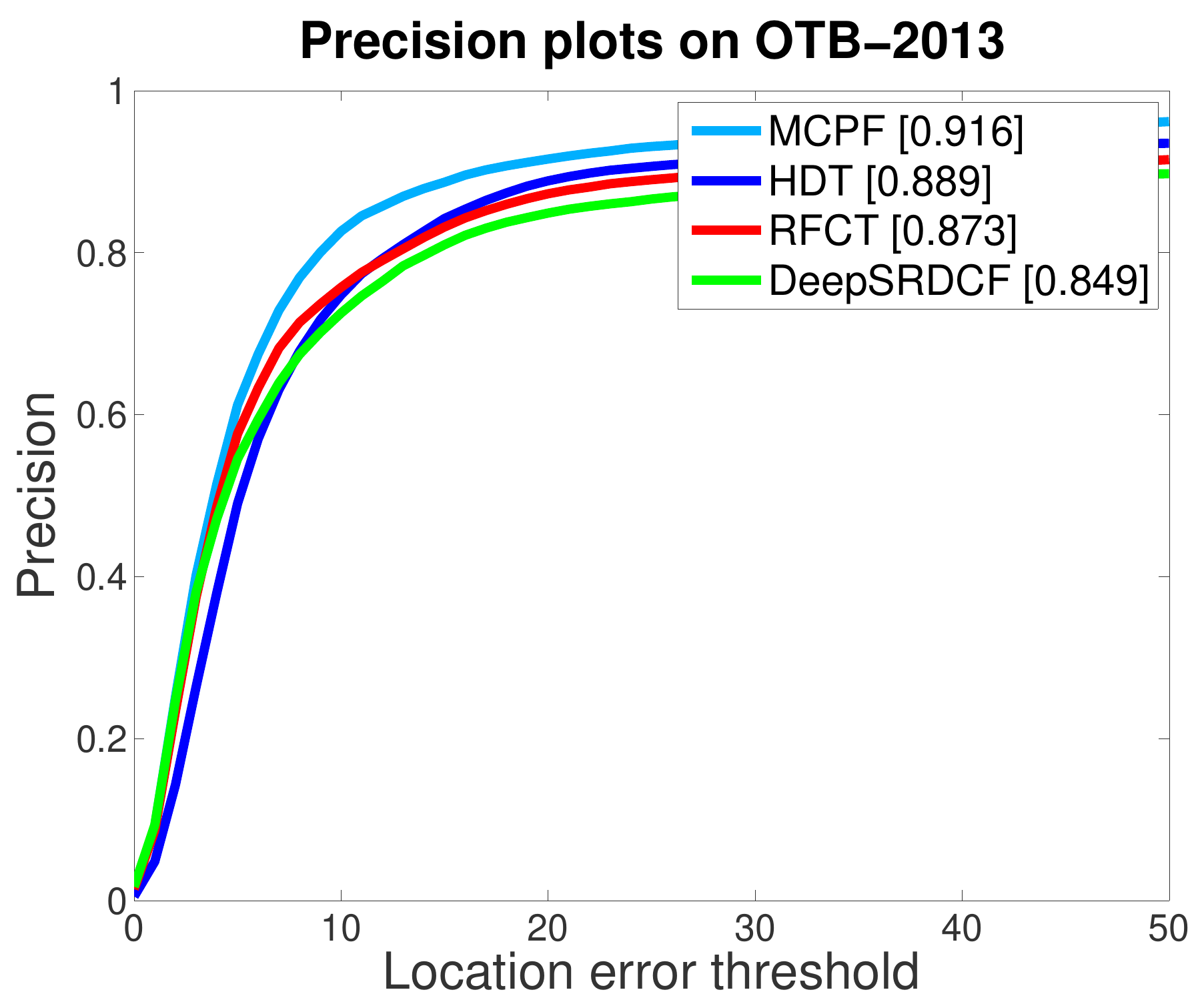}\quad
\includegraphics[width=1.6in]{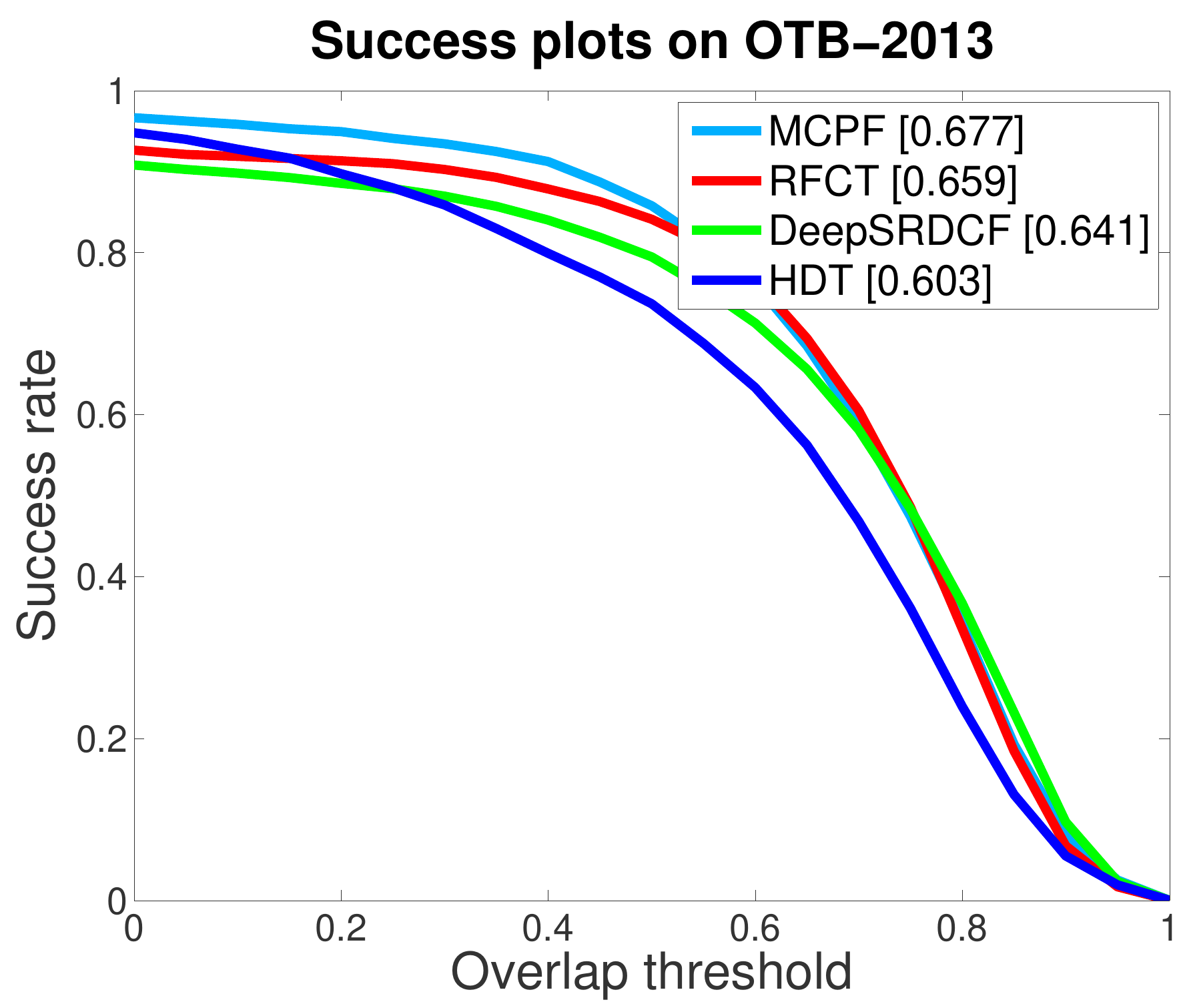}
\includegraphics[width=1.6in]{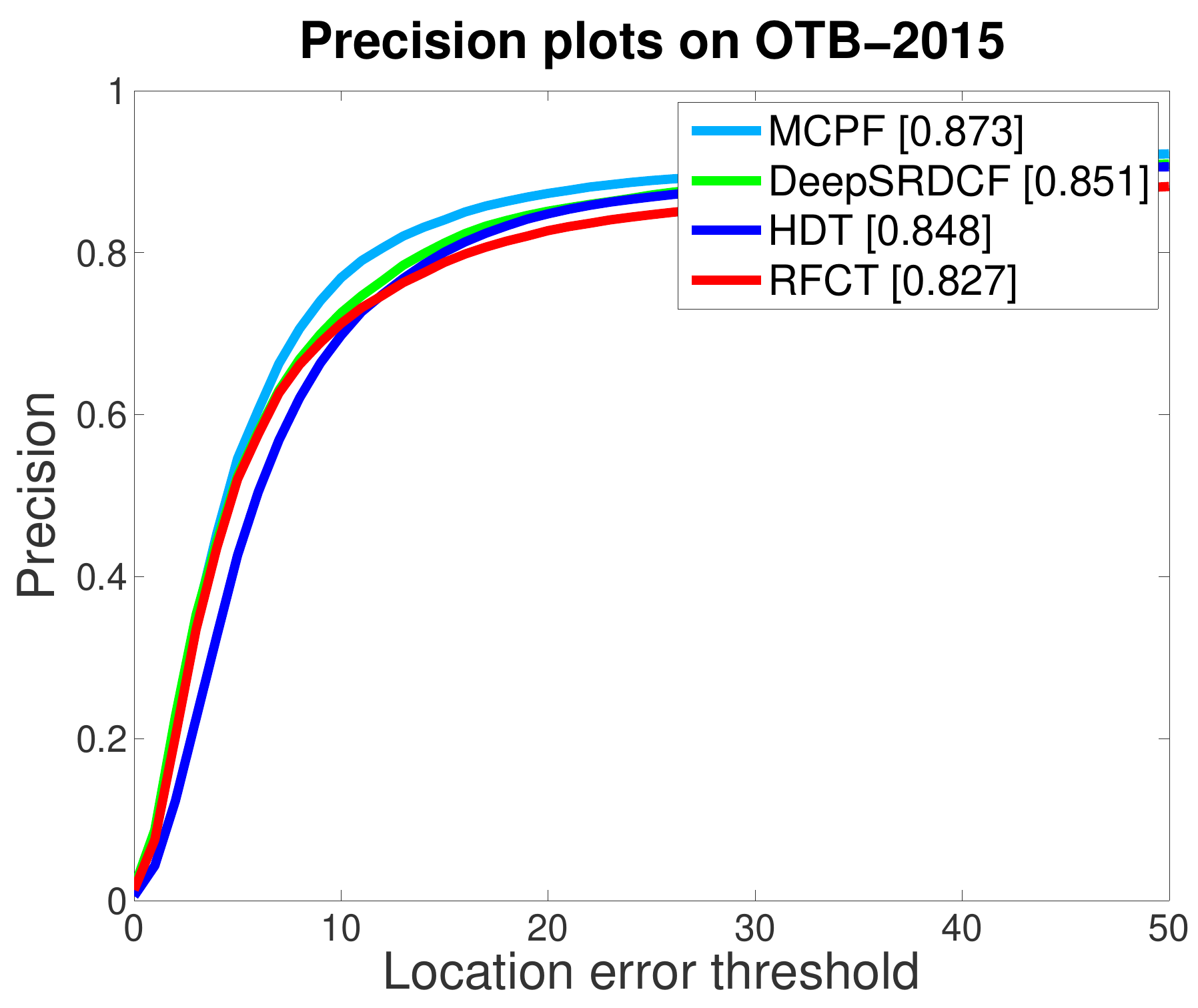}\quad
\includegraphics[width=1.6in]{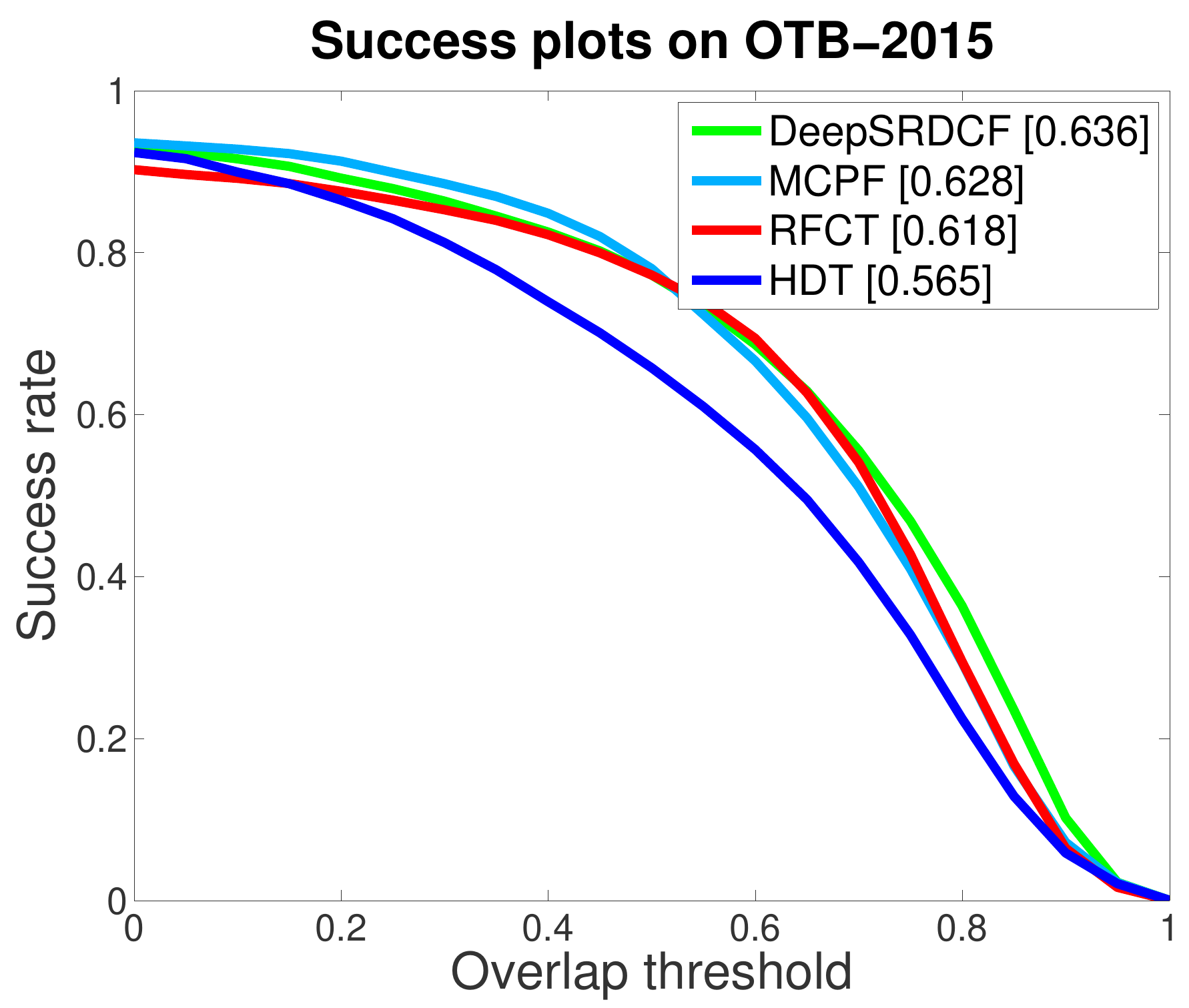}
\caption{Precision and success plots of RFCT compared with the deep features based trackers on OTB-2013 and OTB-2015 datasets. The AUC score of each tracker is displayed in a bracket for success plots.}
\label{Fig:DeepCmp}
\end{figure}

Fig. \ref{Fig:DeepCmp} compares the proposed method based on conventional features with the several deep features based trackers on OTB-2013 and OTB-2015, showing the DP for precision plot and AUC score for success plot of each tracking algorithm. Overall, MCPF performs better among the existing methods. For AUC score, it achieves a gain of 1.8\% on OTB-2013 and 1.0\% on OTB-2015 compared to our approach. Note that, our approach is based on conventional features.
\begin{table*}[htbp]
\vspace{0.5mm}\caption{Success rate (\%) of RFCT compared to the above trackers based on deep features at an overlap threshold 0.5. The \textcolor[rgb]{1,0,0}{first}, \textcolor[rgb]{0,1,0}{second} and \textcolor[rgb]{0,0,1}{third} rank values are highlighted in color.}
%\small
%\renewcommand{\raggedright}{\leftskip=0pt \rightskip=0pt plus 0cm}
\newcommand{\tabincell}[2]{\begin{tabular}{@{}#1@{}}#2\end{tabular}}
 \centering
  \begin{tabular*}{0.8\textwidth}{ @{\extracolsep{\fill}} p{2cm}<{\centering} c c c c}
  \hline
            &\textbf{RFCT} &\tabincell{c}{DeepSRDCF \\ \cite{danelljan2015convolutional}}&\tabincell{c}{HDT \\ \cite{qi2016hedged}} &\tabincell{c}{MCPF \\ \cite{zhang2017multi}}  \\
  \hline
    OTB-2013& \textcolor[rgb]{0,1,0}{84.1}       & \textcolor[rgb]{0,0,1}{79.4} & 73.7       & \textcolor[rgb]{1,0,0}{85.8}  \\
  \hline
    OTB-2015& \textcolor[rgb]{0,1,0}{77.2}       & \textcolor[rgb]{0,1,0}{77.2} &65.8       & \textcolor[rgb]{1,0,0}{78.0} \\
  \hline
 \end{tabular*}
 \label{tab:deepCompThreshold}
\end{table*}

\section{Conclusion}
In this paper, we propose Region-filtering Correlation Tracking (RFCT) to learn a robust filter for visual tracking. Compared with current CF trackers, the proposed RFCT method filters training samples to
focus on more interesting region in training samples by a spatial map. It is a more general way to control background information and target information in training samples. By comparison, Our Map is better than the other two maps in experiments. Moreover, we increase the weight proportions of accurate filters to alleviate model corruption during tracking.
%, and maintains spatial information of training samples. In addition, the values of the spatial map are not restricted, it can be a binary mask map eliminating the influence of IRs, or a map similar to Gaussian penalizing the IRs.
Quantitative evaluations on OTB-2013 and OTB-2015 benchmark datasets demonstrate that the proposed RFCT tracking algorithm performs well against several state-of-the-art trackers.

%\clearpage
%{\small
%\bibliographystyle{ieee}
%\bibliography{egbib}
%}
\bibliographystyle{splncs}

%\bibliographystyle{splncs}
%\bibliography{egbib}

\begin{thebibliography}{10}

\bibitem{danelljan2016eco}
Danelljan, M., Bhat, G., Khan, F.S., Felsberg, M.:
\newblock Eco: Efficient convolution operators for tracking.
\newblock Proceedings of the IEEE Conference on Computer Vision and Pattern
  Recognition (2017)

\bibitem{danelljan2016beyond}
Danelljan, M., Robinson, A., Khan, F.S., Felsberg, M.:
\newblock Beyond correlation filters: Learning continuous convolution operators
  for visual tracking.
\newblock In: Proceedings of European Conference on Computer Vision. (2016)
  472--488

\bibitem{kiani2017learning}
Kiani~Galoogahi, H., Fagg, A., Lucey, S.:
\newblock Learning background-aware correlation filters for visual tracking.
\newblock In: Proceedings of the IEEE Conference on Computer Vision and Pattern
  Recognition. (2017)  1135--1143

\bibitem{mueller2017context}
Mueller, M., Smith, N., Ghanem, B.:
\newblock Context-aware correlation filter tracking.
\newblock (2017)

\bibitem{bolme2010visual}
Bolme, D.S., Beveridge, J.R., Draper, B.A., Lui, Y.M.:
\newblock Visual object tracking using adaptive correlation filters.
\newblock In: Proceedings of the IEEE Conference on Computer Vision and Pattern
  Recognition, IEEE (2010)  2544--2550

\bibitem{henriques2015high}
Henriques, J.F., Caseiro, R., Martins, P., Batista, J.:
\newblock High-speed tracking with kernelized correlation filters.
\newblock IEEE Transactions on Pattern Analysis and Machine Intelligence
  \textbf{37}(3) (2015)  583--596

\bibitem{danelljan2015learning}
Danelljan, M., Hager, G., Shahbaz~Khan, F., Felsberg, M.:
\newblock Learning spatially regularized correlation filters for visual
  tracking.
\newblock In: Proceedings of the IEEE International Conference on Computer
  Vision. (2015)  4310--4318

\bibitem{kiani2015correlation}
Kiani~Galoogahi, H., Sim, T., Lucey, S.:
\newblock Correlation filters with limited boundaries.
\newblock In: Proceedings of the IEEE Conference on Computer Vision and Pattern
  Recognition. (2015)  4630--4638

\bibitem{kiani2013multi}
Kiani~Galoogahi, H., Sim, T., Lucey, S.:
\newblock Multi-channel correlation filters.
\newblock In: Proceedings of the IEEE International Conference on Computer
  Vision. (2013)  3072--3079

\bibitem{danelljan2014adaptive}
Danelljan, M., Shahbaz~Khan, F., Felsberg, M., Van~de Weijer, J.:
\newblock Adaptive color attributes for real-time visual tracking.
\newblock In: Proceedings of the IEEE Conference on Computer Vision and Pattern
  Recognition. (2014)  1090--1097

\bibitem{danelljan2015coloring}
Danelljan, M., H{\"a}ger, G., Khan, F.S., Felsberg, M.:
\newblock Coloring channel representations for visual tracking.
\newblock In: Proceedings of Scandinavian Conference on Image Analysis. (2015)
  117--129

\bibitem{li2014scale}
Li, Y., Zhu, J.:
\newblock A scale adaptive kernel correlation filter tracker with feature
  integration.
\newblock In: Proceedings of European Conference on Computer Vision Workshops.
  (2014)  254--265

\bibitem{tang2015multi}
Tang, M., Feng, J.:
\newblock Multi-kernel correlation filter for visual tracking.
\newblock In: Proceedings of the IEEE International Conference on Computer
  Vision. (2015)  3038--3046

\bibitem{dalal2005histograms}
Dalal, N., Triggs, B.:
\newblock Histograms of oriented gradients for human detection.
\newblock In: Proceedings of the IEEE Computer Society Conference on Computer
  Vision and Pattern Recognition. Volume~1. (2005)  886--893

\bibitem{van2009learning}
Van De~Weijer, J., Schmid, C., Verbeek, J., Larlus, D.:
\newblock Learning color names for real-world applications.
\newblock IEEE Transactions on Image Processing \textbf{18}(7) (2009)
  1512--1523

\bibitem{danelljan2014accurate}
Danelljan, M., H{\"a}ger, G., Khan, F., Felsberg, M.:
\newblock Accurate scale estimation for robust visual tracking.
\newblock In: Proceedings of British Machine Vision Conference. (2014)

\bibitem{bibi2016target}
Bibi, A., Mueller, M., Ghanem, B.:
\newblock Target response adaptation for correlation filter tracking.
\newblock In: Proceedings of European Conference on Computer Vision. (2016)
  419--433

\bibitem{wang2017large}
Wang, M., Liu, Y., Huang, Z.:
\newblock Large margin object tracking with circulant feature maps.
\newblock Proceedings of the IEEE Conference on Computer Vision and Pattern
  Recognition (2017)

\bibitem{hare2016struck}
Hare, S., Golodetz, S., Saffari, A., Vineet, V., Cheng, M.M., Hicks, S.L.,
  Torr, P.H.:
\newblock Struck: Structured output tracking with kernels.
\newblock IEEE transactions on pattern analysis and machine intelligence
  \textbf{38}(10) (2016)  2096--2109

\bibitem{sui2016real}
Sui, Y., Zhang, Z., Wang, G., Tang, Y., Zhang, L.:
\newblock Real-time visual tracking: Promoting the robustness of correlation
  filter learning.
\newblock In: Proceedings of European Conference on Computer Vision. (2016)
  662--678

\bibitem{liu2016structural}
Liu, S., Zhang, T., Cao, X., Xu, C.:
\newblock Structural correlation filter for robust visual tracking.
\newblock In: Proceedings of the IEEE Conference on Computer Vision and Pattern
  Recognition. (2016)  4312--4320

\bibitem{li2015reliable}
Li, Y., Zhu, J., Hoi, S.C.:
\newblock Reliable patch trackers: Robust visual tracking by exploiting
  reliable patches.
\newblock In: Proceedings of the IEEE Conference on Computer Vision and Pattern
  Recognition. (2015)  353--361

\bibitem{liu2015real}
Liu, T., Wang, G., Yang, Q.:
\newblock Real-time part-based visual tracking via adaptive correlation
  filters.
\newblock In: Proceedings of the IEEE Conference on Computer Vision and Pattern
  Recognition. (2015)  4902--4912

\bibitem{zuo2016learning}
Zuo, W., Wu, X., Lin, L., Zhang, L., Yang, M.H.:
\newblock Learning support correlation filters for visual tracking.
\newblock arXiv preprint arXiv:1601.06032 (2016)

\bibitem{zhang2016defense}
Zhang, T., Bibi, A., Ghanem, B.:
\newblock In defense of sparse tracking: Circulant sparse tracker.
\newblock In: Proceedings of the IEEE Conference on Computer Vision and Pattern
  Recognition. (2016)  3880--3888

\bibitem{lukevzivc2016discriminative}
Luke{\v{z}}i{\v{c}}, A., Voj{\'\i}{\v{r}}, T., {\v{C}}ehovin, L., Matas, J.,
  Kristan, M.:
\newblock Discriminative correlation filter with channel and spatial
  reliability.
\newblock Proceedings of the IEEE Conference on Computer Vision and Pattern
  Recognition (2017)

\bibitem{boyd2011distributed}
Boyd, S., Parikh, N., Chu, E., Peleato, B., Eckstein, J.:
\newblock Distributed optimization and statistical learning via the alternating
  direction method of multipliers.
\newblock Foundations and Trends{\textregistered} in Machine Learning
  \textbf{3}(1) (2011)  1--122

\bibitem{wu2013online}
Wu, Y., Lim, J., Yang, M.H.:
\newblock Online object tracking: A benchmark.
\newblock In: Proceedings of the IEEE conference on computer vision and pattern
  recognition. (2013)  2411--2418

\bibitem{wu2015object}
Wu, Y., Lim, J., Yang, M.H.:
\newblock Object tracking benchmark.
\newblock IEEE Transactions on Pattern Analysis and Machine Intelligence
  \textbf{37}(9) (2015)  1834--1848

\bibitem{danelljan2016adaptive}
Danelljan, M., Hager, G., Shahbaz~Khan, F., Felsberg, M.:
\newblock Adaptive decontamination of the training set: A unified formulation
  for discriminative visual tracking.
\newblock In: Proceedings of the IEEE Conference on Computer Vision and Pattern
  Recognition. (2016)  1430--1438

\bibitem{ma2015long}
Ma, C., Yang, X., Zhang, C., Yang, M.H.:
\newblock Long-term correlation tracking.
\newblock In: Proceedings of the IEEE Conference on Computer Vision and Pattern
  Recognition. (2015)  5388--5396

\bibitem{zhang2014meem}
Zhang, J., Ma, S., Sclaroff, S.:
\newblock Meem: robust tracking via multiple experts using entropy
  minimization.
\newblock In: Proceedings of European Conference on Computer Vision. (2014)
  188--203

\bibitem{zhang2017multi}
Zhang, T., Xu, C., Yang, M.H.:
\newblock Multi-task correlation particle filter for robust object tracking.
\newblock In: Proceedings of IEEE Conference on Computer Vision and Pattern
  Recognition. Volume~7. (2017)

\bibitem{danelljan2015convolutional}
Danelljan, M., Hager, G., Shahbaz~Khan, F., Felsberg, M.:
\newblock Convolutional features for correlation filter based visual tracking.
\newblock In: Proceedings of the IEEE International Conference on Computer
  Vision Workshops. (2015)  58--66

\bibitem{qi2016hedged}
Qi, Y., Zhang, S., Qin, L., Yao, H., Huang, Q., Lim, J., Yang, M.H.:
\newblock Hedged deep tracking.
\newblock In: Proceedings of the IEEE Conference on Computer Vision and Pattern
  Recognition. (2016)  4303--4311

\end{thebibliography}

\end{document}